%% file: main.tex
\documentclass[11pt]{article}

\usepackage[final]{acl}

\usepackage{times}
\usepackage{latexsym}

\usepackage[T1]{fontenc}

\usepackage{kotex}

\usepackage{xurl}

\usepackage[utf8]{inputenc}

\usepackage{microtype}

\usepackage{inconsolata}

\usepackage{graphicx}
\usepackage{multirow}
\usepackage{makecell}
\usepackage[normalem]{ulem}
\usepackage{float}
\usepackage{placeins}

\usepackage{hyperref}
\usepackage{url}
\usepackage{xspace}

\usepackage{booktabs}
\usepackage{tabularx}
\usepackage{listings}
\usepackage{listingsutf8}
\usepackage{enumitem}

\usepackage{comment}
\usepackage{titlesec}

\usepackage{soul}
\soulregister\ref{7} %
\soulregister\cite{7} %

\usepackage{xcolor} 
\usepackage{subcaption}
 \usepackage{amsmath} 

\definecolor{codegreen}{rgb}{0,0.6,0}
\definecolor{codegray}{rgb}{0.5,0.5,0.5}
\definecolor{codepurple}{rgb}{0.58,0,0.82}
\definecolor{backcolour}{rgb}{0.95,0.95,0.92}

\lstdefinestyle{mystyle}{
    backgroundcolor=\color{white},
    basicstyle=\ttfamily\scriptsize\color{black},
    commentstyle=\color{black},
    keywordstyle=\color{black},
    stringstyle=\color{black},
    identifierstyle=\color{black},
    numberstyle=\tiny\color{black},
    breaklines=true,
    breakatwhitespace=false,
    keepspaces=true,
    showspaces=false,
    showstringspaces=false,
    showtabs=false,
    tabsize=2,
    frame=none,
    captionpos=b
}
\lstdefinelanguage{yaml}{
  keywords={true,false,null,y,n},
  comment=[l]{\#},
  morecomment=[s]{/*}{*/},
  morestring=[b]",
  morestring=[b]',
  sensitive=true
}

\lstdefinelanguage{python}{
  keywords={def, return, if, else, for, while, break, continue, import, from, as, class, try, except, finally, with, yield, pass, lambda, and, or, not, is, in, True, False, None},
  keywordstyle=\color{magenta},
  ndkeywords={self},
  ndkeywordstyle=\color{codegreen},
  comment=[l]{\#},
  morecomment=[s]{"""}{"""},
  morestring=[b]',
  morestring=[b]",
  sensitive=true
}

\lstdefinelanguage{json}{
    basicstyle=\ttfamily\footnotesize,
    numbers=none,
    showstringspaces=false,
    breaklines=true,
    backgroundcolor=\color{gray!5},
    literate=
     *{0}{{{\color{blue}0}}}{1}
      {1}{{{\color{blue}1}}}{1}
      {2}{{{\color{blue}2}}}{1}
      {3}{{{\color{blue}3}}}{1}
      {4}{{{\color{blue}4}}}{1}
      {5}{{{\color{blue}5}}}{1}
      {6}{{{\color{blue}6}}}{1}
      {7}{{{\color{blue}7}}}{1}
      {8}{{{\color{blue}8}}}{1}
      {9}{{{\color{blue}9}}}{1}
      {:}{{{\color{black}:}}}{1}
      {,}{{{\color{black},}}}{1}
      {"}{{{\color{red}"}}}{1}
      {true}{{{\color{magenta}true}}}{1}
      {false}{{{\color{magenta}false}}}{1}
      {null}{{{\color{magenta}null}}}{1}
}

\newcommand{\bench}{Thunder-KoNUBench\xspace}

\title{Thunder-KoNUBench: A Corpus-Aligned Benchmark\\for Korean Negation Understanding}

\author{
\textbf{Sungmok Jung\thanks{These authors contributed equally to this work.}\textsuperscript{1}}, 
\textbf{Yeonkyoung So\footnotemark[1]\textsuperscript{1}}, 
\textbf{Joonhak Lee\footnotemark[1]\textsuperscript{1}}, \\ 
\textbf{Sangho Kim\textsuperscript{1}},
\textbf{Yelim Ahn\textsuperscript{1}},
\textbf{Jaejin Lee\textsuperscript{1,2}} \\ \\ 
\textsuperscript{1}Graduate School of Data Science, Seoul National University \\
\textsuperscript{2}Dept. of Computer Science and Engineering, Seoul National University \\
\texttt{\{tjdahrwjd,kathy1028,hmjelee,ksh4931,mileya,jaejin\}@snu.ac.kr}\\
\texttt{http://thunder.snu.ac.kr}
}

\begin{document}
\maketitle

\input{latex/0_abstract}
\input{latex/1_intro}
\input{latex/2_related}
\input{latex/3_1_def_neg_tmp}

\input{latex/4_konubench}
\input{latex/5_experiments}
\input{latex/6_conclusion}
\input{latex/limitations}
\input{latex/ethics}
\input{latex/acknowlegments}

\bibliography{custom}
\input{latex/appendix}

\end{document}

%% file: latex/0_abstract.tex
\begin{abstract}
Although negation is known to challenge large language models (LLMs), benchmarks for evaluating negation understanding—especially in Korean—are scarce. We conduct a corpus-based analysis of Korean negation and show that LLM performance degrades under negation. We then introduce \textit{\bench}, a sentence-level negation understanding benchmark that reflects the empirical distribution of Korean negation phenomena. Evaluating 47 LLMs on \bench, we analyze the effects of model size and instruction tuning, and perform error analysis to better understand model behavior. We further show that fine-tuning on \bench improves negation understanding and broader contextual comprehension in Korean
\footnote{Our code and dataset are publicly available on Github and HuggingFace.

\includegraphics[page=2,height=1em]{figures/logo.pdf}
\; \url{https://github.com/mcrl/Thunder-KoNUBench}

\includegraphics[page=1,height=1em,]{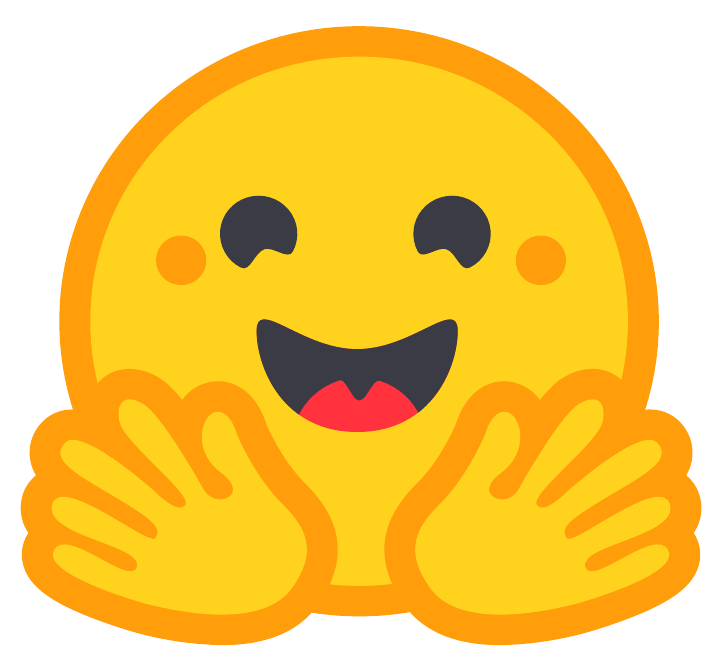}
\; \url{https://huggingface.co/datasets/thunder-research-group/SNU_Thunder-KoNUBench}}.

\end{abstract}

%% file: latex/1_intro.tex
\section{Introduction}
\label{sec:1_introduction}
Negation is a fundamental operation in natural language that reverses the meaning of an expression into its opposite. It enables a variety of linguistic functions, including contradiction, inability, and denial, while also supporting logical reasoning by specifying what is false or absent. Therefore, effectively processing negation is crucial for strong language comprehension. 

However, many studies indicate that large language models (LLMs) often struggle with handling negation~\citep{jumelet2018language, kassner-schutze-2020-negated, truong2023language}. Similar difficulties have been reported in vision-language models (VLMs)~\citep{alhamoud2025vision, park2025know}. In response to these challenges, several datasets~\citep{garcia2023not, hossain2022question, hartmann2021multilingual, ravichander2022condaqa} and training strategies~\citep{hosseini2021understanding, singh2023nlms, han2025negation, park2025know} have been proposed to enhance or evaluate negation handling in both LLMs and VLMs. Unfortunately, these efforts have predominantly focused on English. Despite the development of recent benchmarks for negation phenomena in languages such as Czech, German, Ukrainian, Bulgarian, and French~\citep{hartmann2021multilingual, vrabcova2025negation}, there is a notable lack of benchmarks and evaluation studies concerning negation in Korean. Consequently, little is known about whether similar limitations exist in Korean, and no benchmark currently exists that systematically evaluates sentence-level negation understanding in Korean.

To address this issue, we first analyze the characteristics and distribution of negation in Korean and examine whether existing LLMs suffer performance degradation in the presence of negation. We then introduce \bench, a multiple-choice benchmark that reflects the empirical distribution of Korean negation phenomena. Finally, we evaluate a wide range of Korean and non-Korean LLMs on \bench and investigate fine-tuning strategies for improving sentence-level negation understanding.

The main contributions of this paper are summarized as follows: 
\begin{itemize}[leftmargin=6mm] \setlength{\itemsep}{0pt} \setlength{\parskip}{0pt} 
\item We demonstrate that LLMs, both Korean and non-Korean, encounter difficulties when handling negation in Korean, even in simple sentence-level tasks. They experience significant performance degradation when required to reason with negation. 
\item We conduct a comprehensive corpus study of Korean negation, analyzing the statistical distribution of major negation types and the sentence structures in which they are used. 
\item We introduce \textit{\bench}, a sentence-level multiple-choice benchmark that systematically reflects the empirical distribution of Korean negation phenomena. \item We evaluate 47 LLMs (18 Korean models and 29 non-Korean models) using \bench to analyze the effects of model size and instruction tuning on negation understanding. 
\item Our experiments indicate that supervised fine-tuning on \bench enhances the models’ contextual understanding in Korean, and that cloze-style supervision is more effective than symbol-style supervision for learning sentence-level negation. \end{itemize}

%% file: latex/2_related.tex
\section{Related Work}
\subsection{Challenges of LLMs in Negation Handling} 
Numerous studies have demonstrated that negation presents considerable challenges for language models. Researches indicate that pretrained language models (PLMs) often struggle to differentiate between negated and non-negated questions~\citep{kassner-schutze-2020-negated, ettinger2020bert}. These findings have been questioned as to whether the observed performance degradation truly reflects an inability to handle negation, or rather stems from difficulties in contextual or factual reasoning~\citep{gubelmann2022context}. However, subsequent work using more fine-grained evaluation settings shows that some PLMs still struggle to handle negation robustly under controlled conditions\citep{kletz2023self}. Furthermore, their inability to understand synonym-antonym relationships contributes to failures in reasoning under negation~\citep{truong2023language}. 

These shortcomings have been identified as key factors leading to degraded performance across various downstream NLP tasks, including machine translation~\citep{hossain2020s, tang2021revisiting}, information extraction~\citep{grivas2020not}, and sentiment analysis~\citep{barnes2021improving}. More recently, similar findings have been observed in multimodal language models, such as VLMs, where improved understanding of negation significantly impacts performance~\citep{alhamoud2025vision}. Enhancing awareness of negation has also been shown to yield measurable improvements~\citep{park2025know}.

\subsection{Datasets and Benchmarks for Negation} 
Previous studies have highlighted that both NLP corpora and downstream tasks contain very few instances of negation~\citep{hossain-etal-2020-analysis, hossain2022analysis}. It has led to an increasing focus on benchmarks specifically designed to evaluate negation comprehension. NaN-NLI~\citep{truong2022not} and ScoNe~\citep{ravichander2022condaqa} assess whether models can accurately determine premise–hypothesis relationships in the presence of negation. Additionally, CONDAQA~\citep{ravichander2022condaqa} evaluates whether models understand the semantic implications of negated statements. Thunder-NUBench~\citep{so2026thunder} tests if models can identify the correct standard negation of an original sentence, with a clearly defined scope of negation. However, these efforts are primarily focused on English and a few high-resource languages. 

Although multilingual benchmarks have been proposed~\citep{hartmann2021multilingual, vrabcova2025negation}, there are no comparable resources or systematic studies of negation in low-resource languages such as Korean. Among existing Korean benchmarks, KoBest~\citep{jang-etal-2022-kobest} introduces SentiNeg, a sentiment analysis task involving negation, but its coverage of negation is limited, and the sentences are overly simplistic. KMMLU~\citep{son2025kmmlu} includes a subset of questions that contain negation, yet these items are mostly easy declarative questions, failing to capture the impact of negation on model performance meaningfully. To address this gap, we examine the impact of negation in Korean and introduce a benchmark specifically targeting negation understanding, carefully designed to reflect the linguistic characteristics of Korean negation.

%% file: latex/3_1_def_neg_tmp.tex
\section{Korean Negation}
Negation in Korean is expressed through various morphological, syntactic, and semantic mechanisms. Despite its complexity, there has been limited empirical investigation into its statistical properties, and few analyses have explored its impact on the performance of LLMs. 
In this section, we define the negation in Korean, analyze their statistical distribution, and evaluate LLMs' robustness to Korean negation.



\subsection{Definition of Negation in Korean}
Logically, negation is an operation that reverses the meaning of an original sentence. If a sentence expresses the proposition $P$, its logical negation corresponds to $\lnot P$. This complementary relationship is distinct from contradiction, which simply refers to the incompatibility between two propositions. To implement logical negation, Thunder-NUBench~\citep{so2026thunder} introduces standard negation as a logical operation that reverses the truth value of the entire sentence. This is different from local negation, which provides only a partial negation of the overall meaning of the sentence. In this work, while adopting the core concepts of Thunder-NUBench, we redefine standard negation and local negation to systematically reflect the linguistic properties of Korean.

\paragraph{Standard Negation.}
Standard negation is defined as a recursive operation applied to the logical structure among main clauses. It first negates the logical relationship (e.g., conjunction, disjunction, implication) connecting the main clauses and is then recursively applied to each main clause. This process continues until only atomic propositions remain, at which point standard negation is realized by negating the predicate of each atomic proposition. This concept is grounded in two linguistic assumptions:
(1) a main clause can express a complete meaning on its own and thus functions as an atomic proposition, and
(2) the predicate serves as the head of the clause~\citep{miller2011critical}.

When negating predicates, diverse negative markers in Korean may be used. Korean employs a variety of negative markers such as “안”, “못”, “-지 않-”, “-지 못하-”, and “-지 말-”. Each of these markers conveys a distinct semantic nuance and is subject to specific syntactic and semantic constraints on its usage (see Appendix~\ref{app:neg_korean} for a detailed linguistic description of Korean negation). 
When negating the predicate of a clause, these markers may be used depending on the predicate type and sentence context.

\input{tables/4_1_local_neg_typology}

\paragraph{Local Negation.} 
Local negation refers to negation applied locally, producing a partial negation that does not reverse the  meaning of original sentence completely. This includes cases where negation targets a dependent clause or only one of multiple main clauses. In Thunder-NUBench~\citep{so2026thunder}, the classification of local negation based on English sentence structure is divided into several categories: negation in relative clauses, participial clauses, adverbial clauses, and compound sentences with local negation. 

In this paper, we redefine the typology of local negation in terms of Korean sentence structure. Clause combinations in Korean take diverse structural forms. Korean clauses may be linked via connective endings or conjunctive particles, or embedded within another clause as part of a larger syntactic structure (see Appendix~\ref{app:korean_sentence} for a detailed discussion of Korean sentence structure). Our classification includes negation in noun clauses, adnominal clauses, adverbial clauses, quotation clauses, subordinate clauses, and coordinated sentences with local negation. Examples of each category are provided in Table~\ref{tab:local_negation}.

\input{tables/3_1_corpus}

\subsection{Statistics of Korean Negation}
\label{subsec:3_2_dists}
We analyze the statistical distribution of negation in Korean using a large-scale corpus from the OpenAI Dataset Project~\citep{AiHub}. The dataset we utilize is titled "한국어 성능이 개선된 초거대 AI 언어모델 개발 및 데이터" (Dataset and Large Language Models with Improved Korean Performance). 
It includes both spoken and written-style texts and covers a wide range of topics, reflecting diverse real-world language use. Further details on the corpus are provided in Appendix~\ref{app:corpus}.

To begin the analysis, we segment all sentences in the corpus using \texttt{split\_sentences} function in the \texttt{KSS} library. We then randomly sample 30,000 sentences using a fixed random seed 2025. Upon inspection, we find that some sentences are excessively long to be treated as single sentences, so we remove sentences longer than 400 characters, resulting in a total of 29,476 sentences. Next, we apply a rule-based method to identify sentences containing negation. Then, the authors manually verify the candidates to ensure they accurately represent genuine instances of negation.

Through this verification process, we identify 3,160 negative sentences, indicating that approximately 10.7\% of the Korean corpus consists of negative sentences. Furthermore, we examine the statistical distribution of these negations. Table~\ref{tab:corpus1} presents the distribution of each type of negation in the corpus, along with the types of sentences in which each negation expression appears. For a detailed linguistic description of Korean sentence structure, please see Appendix~\ref{app:korean_sentence}.

\begin{figure*}[!tb]
\centering
\begin{footnotesize}

\begin{subfigure}[t]{\linewidth}
    \centering
    \includegraphics[page=1,width=\linewidth]{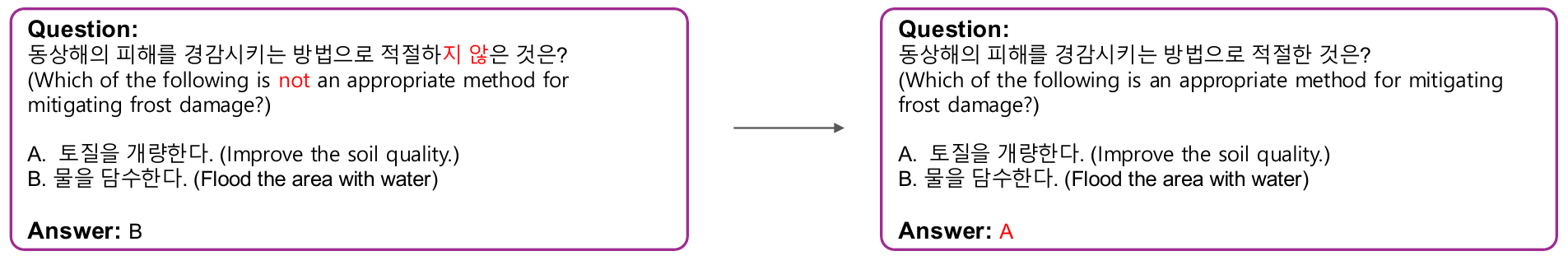}
    \caption{An example of binary-format KMMLU questions containing negation (left) and their affirmative counterparts (right).}
    \label{fig:kmmlu_boolq_a}
\end{subfigure}

\vspace{\baselineskip}

\begin{subfigure}[t]{\linewidth}
    \centering
    \includegraphics[page=2,width=\linewidth]{figures/kmmlu_boolq.pdf}
    \caption{An example of KoBest BoolQ questions in their original (left) and negated (right) forms.}
    \label{fig:kmmlu_boolq_b}
\end{subfigure}

\end{footnotesize}
\vspace*{-0.5\baselineskip}
\caption{Illustration of negation-induced performance evaluation on KMMLU and KoBest BoolQ.}
\label{fig:kmmlu_boolq}
\vspace{0.5\baselineskip}
\end{figure*}

\begin{figure*}[!t]
\centering
\includegraphics[width=\linewidth]{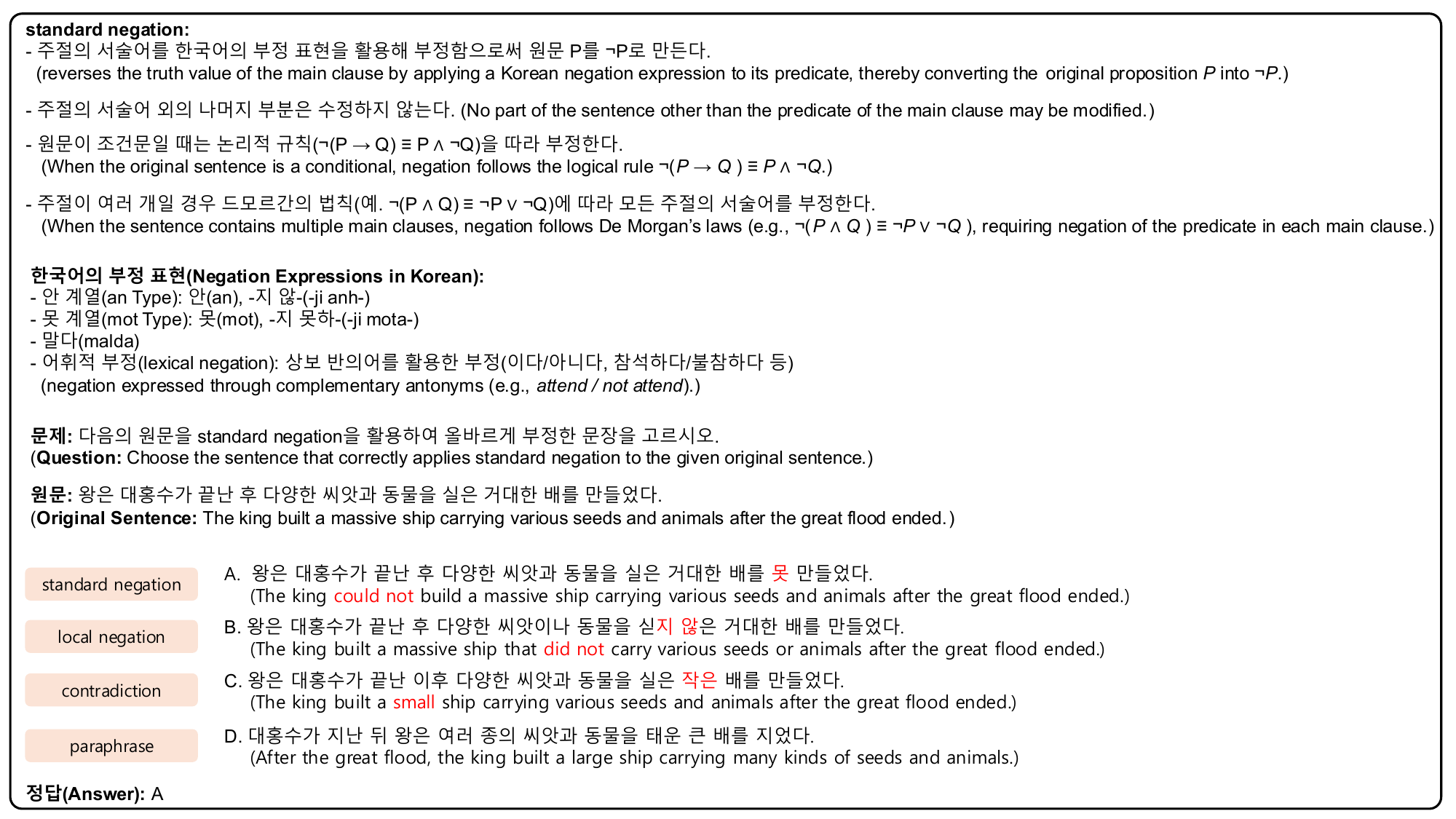}
\vspace*{-1.5\baselineskip}
\caption{An example instance from \bench.}
\label{fig:example}
\end{figure*}

\subsection{LLMs on Korean negation}
\label{sec:performance}

\input{tables/3_2_kmmlu_boolq}

\paragraph{LLM performance on KMMLU.} 

KMMLU is a multiple-choice dataset featuring 45 categories that reflect Korea's cultural and regional characteristics. While the benchmark includes questions involving negation, it does not sufficiently address whether language models genuinely struggle with negation in Korean. In fact, the authors pointed out that models tend to perform better on items with negation compared to items without negation in the KMMLU test set. Their analysis suggests that this result arises because the negative questions in KMMLU focus primarily on relatively simple declarative knowledge, rather than procedural knowledge~\citep{son2025kmmlu}.

To investigate whether models face difficulties with negation even in simple declarative questions, we conducted an additional experiment. First, we extracted a total of 7,153 questions containing negation from KMMLU using a rule-based detector and converted them into binary-choice items for evaluation. Next, we transformed these items into affirmative sentences and re-evaluated the models. Figure~\ref{fig:kmmlu_boolq_a} provides an example of this process. The results, shown in Table~\ref{tab:kmmlu_boolq_brief}, indicate that both Korean and non-Korean models achieve higher performance on the affirmative versions on average. These findings suggest that models struggle to handle negation, even in simple declarative questions.

\paragraph{LLM performance on KoBest BoolQ.} 
KoBest BoolQ is a binary-choice dataset featured in KoBest~\citep{jang-etal-2022-kobest}, in which models are required to respond with either \textit{True} or \textit{False} based on their understanding of the provided context. This task is more challenging than the straightforward declarative questions in KMMLU, as it requires context-sensitive reasoning. To investigate the impact of negation, we transformed all 1,404 questions into their negated forms and assessed the models' performance. An example of this process is shown in Figure~\ref{fig:kmmlu_boolq_b}. The results indicate a significant decline in performance across all 43 models, reinforcing the previous finding that LLMs struggle to reason under negation~\citep{truong2023language}, and this issue is also relevant in Korean.

%% file: tables/4_1_local_neg_typology.tex
\begin{table}[!t]
\centering
\footnotesize
\resizebox{0.9\linewidth}{!}{%
\setlength{\tabcolsep}{4pt} 
\begin{tabularx}{\linewidth}{@{}
  >{\raggedright\arraybackslash}p{0.18\linewidth}
  >{\centering\arraybackslash}X
@{}}
\toprule
\textbf{Category} & \textbf{Example} \\
\midrule\midrule

\textbf{Noun clauses Negation} &
농부들이 비가 오기를 기다린다. → 농부들이 비가 오{\color{red}지 않}기를 기다린다. \newline
(The farmers are waiting for it to rain. → The farmers are waiting for it {\color{red}not} to rain.) \\\midrule

\textbf{Adnominal clauses Negation} &
내가 태어난 1950년에 전쟁이 발발했다. → 내가 태어나{\color{red}지 않}은 1950년에 전쟁이 발발했다. \newline
(In 1950, the year I was born, a war broke out. → In 1950, the year I was {\color{red}not} born, a war broke out.) \\\midrule

\textbf{Quotation clauses Negation} &
나는 그가 나의 연설에 만족했다고 들었다. → 나는 그가 나의 연설에 만족하{\color{red}지 않}았다고 들었다. \newline
(I heard that he was satisfied with my speech. → I heard that he was {\color{red}not} satisfied with my speech.) \\\midrule

\textbf{Subordinate clauses Negation} &
꽃이 잘 자라도록 나는 물을 주었다. → 꽃이 잘 자라{\color{red}지 못}하도록 나는 물을 주었다. \newline
(I watered the flowers so that they would grow well. → I watered the flowers so that they {\color{red}would not} grow well.) \\\midrule

\textbf{Adverbial clauses Negation} &
우리는 그녀가 지나가도록 길을 비켜 주었다. → 우리는 그녀가 지나가{\color{red}지 못}하도록 길을 비켜 주었다. \newline
(We stepped aside to let her pass. → We stepped aside so that she {\color{red}could not} pass.) \\\midrule
\textbf{Coordinated sentence with local negation} &
나는 밥을 먹었고, 친구는 빵을 먹었다. → 나는 밥을 {\color{red}안} 먹었고, 친구는 빵을 먹었다. \newline
(I ate rice, and my friend ate bread. → I {\color{red}didn’t} eat rice, and my friend ate bread.) \\

\bottomrule
\end{tabularx}
}
\vspace{-0.5\baselineskip}
\caption{Examples of Local Negation Typology in Korean.}
\label{tab:local_negation}
\end{table}

%% file: tables/3_1_corpus.tex
\begin{table*}[!t]
\centering
\resizebox{0.85\linewidth}{!}{%
\setlength{\tabcolsep}{2pt} 
\begin{tabular}{@{}ccccccc@{}}
\toprule
\multirow{2}{*}{} 
& \multicolumn{2}{c}{\textbf{안 계열(\textit{an} type)}}               
& \multicolumn{2}{c}{\textbf{못 계열(\textit{mot} type)}}               
& \multirow{2}{*}{\textbf{\makecell{말다(\textit{malda})}}} 
& \multirow{2}{*}{\textbf{Total}} 
\\ \cmidrule(lr){2-3} \cmidrule(lr){4-5}
                  
  & \multicolumn{1}{c}{\textbf{Short-Form}} & \textbf{Long-Form} 
  & \multicolumn{1}{c}{\textbf{Short-Form}} & \textbf{Long-Form} 
  &                             
  &                      
  \\ \midrule
        \textbf{\makecell{Instance (\%)}}          
        & \makecell{795 (25.16\%)} 
        & \makecell{1,598 (50.57\%)} 
        & \makecell{174 (5.51\%)} 
        & \makecell{473 (14.97\%)} 
        & \makecell{120 (3.8\%)}                         
        & \makecell{3,160 (100\%)} 


\\ \bottomrule
\end{tabular}%
}

\vspace{0.5em}

\resizebox{0.85\linewidth}{!}{%
\setlength{\tabcolsep}{2pt} 
\begin{tabular}{@{}cccccccc@{}}
\toprule
& \multicolumn{1}{c}{\textbf{\begin{tabular}[c]{@{}c@{}}Adnominal
\\Clause\end{tabular}}}
& \multicolumn{1}{c}{\textbf{\begin{tabular}[c]{@{}c@{}}Adverbial \\Clause\end{tabular}}} 
& \multicolumn{1}{c}{\textbf{\begin{tabular}[c]{@{}c@{}}Noun \\Clause\end{tabular}}} 
& \multicolumn{1}{c}{\textbf{\begin{tabular}[c]{@{}c@{}}Quotation \\Clause\end{tabular}}} 
& \multicolumn{1}{c}{\textbf{\begin{tabular}[c]{@{}c@{}}Subordinate \\Clause\end{tabular}}} 
&  \multicolumn{1}{c}{\textbf{\begin{tabular}[c]{@{}c@{}}Main \\Clause\end{tabular}}} 
& \textbf{Total}  
\\ \midrule
    \textbf{\makecell{Instance (\%)}}        
    & \makecell{1,022 (32.34\%)}             
    & \makecell{624  (19.75\%)}              
    & \makecell{98  (3.1\%)}          
    & \makecell{132  (4.18\%)}              
    & \makecell{241 (7.63\%)}              
    & \makecell{1,043  (33.01\%)}        
    & \makecell{3,160 (100\%)} 
\\ \bottomrule
\end{tabular}%
}
\vspace{-0.5\baselineskip}
\caption{Distribution of syntactic negation in Korean. The upper part shows the distribution of negation types, while the lower part presents the distribution of clause types in which negation appears.}
\label{tab:corpus1}
\end{table*}

%% file: tables/3_2_kmmlu_boolq.tex
\begin{table}[!t]
\centering
\setlength{\tabcolsep}{2pt} 
\resizebox{\columnwidth}{!}{%
\begin{tabular}{lcccc}
\toprule
& \multicolumn{2}{c}{\textbf{KMMLU}} 
& \multicolumn{2}{c}{\textbf{BoolQ}} \\
\cmidrule(lr){2-3} \cmidrule(lr){4-5}
& \textbf{Negative} & \textbf{Affirmative}
& \textbf{Original} & \textbf{Negated} \\
\midrule
\textbf{Korean Models}      & 63.0 & 66.0 & 73.2 & 58.6 \\
\textbf{Non-Korean Models}  & 62.7 & 63.6 & 62.8 & 50.2 \\
\textbf{All Models}         & 62.8 & 64.6 & 67.2 & 53.7 \\
\bottomrule
\end{tabular}
}
\vspace{-0.5\baselineskip}
\caption{Average performance of models on KMMLU and BoolQ with and without negation.}
\label{tab:kmmlu_boolq_brief}
\end{table}

%% file: latex/4_konubench.tex
\input{tables/4_1_category}


\section{\bench}

\subsection{Task Overview}
\bench is a multiple-choice dataset consisting of 4,784 instances, designed to evaluate sentence-level understanding of negation in Korean (see Figure~\ref{fig:example} for an example). In terms of the scope and categorization of negation, \bench closely follows the structure of Thunder-NUBench~\citep{so2026thunder}, while appropriately adapting it to reflect the linguistic properties of the Korean language. Specifically, models are required to select the correct standard negation of the original sentence from various distractors, which include local negation, contradiction, and paraphrase. For a detailed description of each category, see Table \ref{tab:category}. To choose the correct answer, models must identify the main clause and its primary predicate and then apply the appropriate negation to it.

\input{tables/4_2_statistics}

\paragraph{Distributions of negation.} 
Table~\ref{tab:stat1} shows the distribution of negation in \bench. We analyze the types of negation present in both standard and local contexts, as well as the clause types in which local negation occurs. Since the KL divergence from the distribution observed in the corpus (Table~\ref{tab:corpus1}) is very small, we conclude that our benchmark closely aligns with the distributional properties of negation in Korean.

\subsection{Construction of \bench}
The \bench dataset is developed through three main stages: (1) pre-processing of original sentences, (2) generating each choice, and (3) review. Detailed prompt examples used in the construction process are provided in Appendix~\ref{app:construct}.

\paragraph{Pre-processing of original sentences.}  
We crawl the Korean Wikipedia~\citep{Wikipedia_Daemun_2025} and split the text into pairs of sentences using a rule-based method. Then, we employ the OpenAI API~\citep{OpenAI_2025} to merge each pair into a single, well-formed sentence. This approach of splitting the text into two-sentence units before merging them is designed to avoid overly simplistic original sentences. The authors conduct a manual verification process to correct any grammatical errors and unnatural expressions, thereby finalizing the set of original sentences.

\paragraph{Generation of choices.}
Each original sentence is paired with four options: standard negation, local negation, contradiction, and paraphrase. Descriptions of these options are provided in Table~\ref{tab:category}. These four options serve as candidates for the negated form of the original sentence, and the model must select the standard negation by following the instructions and demonstrating an understanding of sentence structure and Korean negation expressions. Standard negation and local negation are created manually by the authors, without the use of language models. When language models are employed, the generated sentences often fail to negate the intended part correctly or introduce unintended modifications elsewhere. Since \bench is designed to assess whether a model can accurately identify the main clause's predicate and negate it appropriately, the authors manually construct these two categories. In contrast, the contradiction and paraphrase options are initially generated using the OpenAI API~\citep{OpenAI_2025} and are subsequently refined through a thorough review process conducted by the authors.

\paragraph{Review process.}
The construction of the original sentences and all response options undergoes thorough cross-checking among the authors. For each item, a different author, independent of the creator, verifies its correctness. Any disagreements are addressed collectively during regular meetings (see Appendix~\ref{app:review} for details on the review process and inter-annotator agreement). This process ensures that all authors reach a consensus on every item, which upholds the quality of the final dataset. Table~\ref{tab:split_stat} displays the statistics of the final dataset.

\subsection{Human Evaluation}
After creating the dataset, we conduct a human evaluation to verify its reliability and establish a baseline for human performance. We recruit ten participants who are not part of our research group and ask them to solve 50 questions randomly selected from the test set. To ensure the evaluation's validity, we restrict all participants' internet access to prevent the use of external resources.

The evaluation is conducted by asking participants to select the option they consider correct for each question. Performance is measured using standard accuracy; since the test consists of 50 questions, each correct answer is assigned 2 points (and 0 points for incorrect answers), resulting in a total score out of 100. The human performance results on \bench are presented in Table~\ref{tab:human_eval}. The consistently high human performance suggests that our benchmark is internally consistent and linguistically sound, serving as a reliable testbed for model evaluation.

%% file: tables/4_1_category.tex
\begin{table*}[htbp]
\centering
\resizebox{\textwidth}{!}{%
\begin{tabular}{lp{16cm}}
\hline
\textbf{Category} & \textbf{Description} \\ \hline\hline
\textbf{Standard Negation} 
& Sentences that negate the main predicate of the main clause, thereby reversing the truth value of the original sentence. The main predicate is negated using Korean syntactic negation or complementary antonyms
. 
This category constitutes the correct answer choice in \bench.\\\hline
\textbf{Local Negation} 
& Sentences in which negation applies locally—either to a predicate in a dependent clause or to only one of multiple main clauses—thereby producing a partial negation that does not reverse the overall sentence meaning.
\\\hline
\textbf{Contradiction} 
& Sentences whose meaning is incompatible with the original sentence, but which do not use syntactic or lexical negation. The meaning is altered through gradable antonyms, different numerical values, changes of entity, or other semantic shifts.
\\\hline
\textbf{Paraphrase}
& Sentences that preserve the meaning of the original sentence while changing its surface form. This may involve using synonyms or altering the syntactic structure, resulting in sentences that must be true whenever the original sentence is true.\\\hline
\end{tabular}
}
\vspace{-0.5\baselineskip}
\caption{Multiple choice categories included in \bench.}
\label{tab:category}
\end{table*}

%% file: tables/4_2_statistics.tex
\begin{table*}[htbp]
\centering
\setlength{\tabcolsep}{3pt} 
\resizebox{0.9\textwidth}{!}{%
\begin{tabular}{@{}cccccccc@{}}
\toprule
               & \multicolumn{2}{c}{\textbf{안 계열(\textit{an} type)}}               
               & \multicolumn{2}{c}{\textbf{못 계열(\textit{mot} type)}}               
               & \multirow{2}{*}{\textbf{말다(\textit{malda})}} 
               & \multirow{2}{*}{\textbf{Total}} 
               & \multirow{2}{*}{\textbf{$D_{\mathrm{KL}}(P \| Q)$}}
               \\ \cmidrule(r){2-3} \cmidrule{4-5}
               & \multicolumn{1}{c}{\textbf{Short-Form}} & \textbf{Long-Form} 
               & \multicolumn{1}{c}{\textbf{Short-Form}} & \textbf{Long-Form} 
               &                            
               &                      
               \\ \midrule
\textbf{\makecell{Instance(\%)}}       
& \multicolumn{1}{c}{\makecell{1,991 (20.92\%)}}       
& \makecell{5,377 (56.49\%)}      
& \multicolumn{1}{c}{\makecell{671 (7.05\%)}}        
& \makecell{1,210 (12.71\%)}      
& \makecell{269 (2.83\%)}                         
& \makecell{9,518 (100\%)}                  
& 0.007430
\\ \bottomrule
\end{tabular}%
}

\vspace{0.5em}

\resizebox{0.9\textwidth}{!}{%
\begin{tabular}{@{}ccccccccc@{}}
\toprule
               & \multicolumn{1}{c}{\textbf{\begin{tabular}[c]{@{}c@{}}Adnominal \\Clause\end{tabular}}}
               & \multicolumn{1}{c}{\textbf{\begin{tabular}[c]{@{}c@{}}Adverbial \\Clause\end{tabular}}} 
               & \multicolumn{1}{c}{\textbf{\begin{tabular}[c]{@{}c@{}}Noun \\Clause\end{tabular}}} 
               & \multicolumn{1}{c}{\textbf{\begin{tabular}[c]{@{}c@{}}Quotation \\Clause\end{tabular}}} 
               & \multicolumn{1}{c}{\textbf{\begin{tabular}[c]{@{}c@{}}Subordinate \\Clause\end{tabular}}} 
               &  \multicolumn{1}{c}{\textbf{\begin{tabular}[c]{@{}c@{}}Coordinated \\Sentence\end{tabular}}} 
               & \textbf{Total}
               & \textbf{$D_{\mathrm{KL}}(P \| Q)$}
               \\ \midrule
\textbf{\makecell{Instance(\%)}}       
& \makecell{1,433 (33.73\%)}
& \makecell{891 (20.97\%)}
& \makecell{134 (3.15\%)}
& \makecell{152 (3.58\%)}
& \makecell{243 (5.72\%)}
& \makecell{1,395 (32.84\%)}
& \makecell{4,248 (100\%)}
& 0.413696
\\ \bottomrule
\end{tabular}%
}
\vspace{-0.5\baselineskip}
\caption{Distribution of negation in \bench. The upper part shows the distribution of negation types, while the lower part presents the distribution of clause types in which negation appears under local negation. The KL divergence is computed as $D_{\mathrm{KL}}(P \| Q)$, where P denotes the distribution observed in the Korean corpus in Section~\ref{subsec:3_2_dists} and Q denotes the distribution in \bench.}
\label{tab:stat1}
\end{table*}

\begin{table}[!t]
\centering
\begin{minipage}{\linewidth}
\centering
\resizebox{0.7\columnwidth}{!}{%
\begin{tabular}{@{}ccccc@{}}
\toprule
\textbf{Split} 
& \textbf{Train} 
& \textbf{Validation} 
& \textbf{Test} 
& \textbf{Total} 
\\ \midrule
\textbf{Count}      
& 2,500 
& 1,000
& 1,284
& 4,784
\\ \bottomrule
\end{tabular}%
}
\vspace{-0.5\baselineskip}
\caption{\bench statistics.}
\label{tab:split_stat}
\end{minipage}\\

\vspace{0.5\baselineskip}

\begin{minipage}{\linewidth}
\centering
\resizebox{0.55\columnwidth}{!}{%
\begin{tabular}{@{}cccc@{}}
\toprule
\textbf{Max} 
& \textbf{Min} 
& \textbf{Median} 
& \textbf{Average} 
\\ \midrule
100.0      
& 88.0 
& 98.0
& 97.6
\\ \bottomrule
\end{tabular}%
}
\vspace{-0.5\baselineskip}
\caption{Human evaluation results on \bench.}
\label{tab:human_eval}
\end{minipage}
\vspace{-1\baselineskip}
\end{table}

%% file: latex/5_experiments.tex
\section{Experiments}

\subsection{Experimental Setup}
\paragraph{Models.}
We conduct experiments on 47 large language models (LLMs), including several variants of Qwen 3~\citep{yang2025qwen3}, Mistral~\citep{jiang2023mistral7b}, Llama 3.1 and Llama 3.2~\citep{grattafiori2024llama3herdmodels}, GPT-4.1~\citep{achiam2023gpt}, and Claude-Opus 4.5~\citep{ANTHROPIC_2025}. Our evaluation also includes nearly all major Korean LLMs released to date, such as Kanana 1.5~\citep{bak2025kanana}, mi:dm 2.0~\citep{KT_2025}, hyperclobaX~\citep{naver2025hyperclova}, EXAONE-4.0~\citep{research2025exaone1}, EXAONE-Deep~\citep{research2025exaone2}, and A.X 4.0~\citep{SKT_2025}, along with many widely used non-Korean models. More details can be found in Appendix~\ref{app:models}. This comprehensive selection ensures that our results accurately reflect the performance of both domestic and international state-of-the-art LLMs.

\paragraph{Evaluation method.}
We evaluate models using the LM Evaluation Harness~\citep{sutawika2025eleutherai} in two standard settings for Multiple-Choice Question Answering (MCQA): \textit{cloze} and \textit{symbol}. In the cloze setting, when presented with a question, the model determines which option has the highest log-likelihood. We report performance using length-normalized accuracy (\textit{acc\_norm}), calculated by dividing the raw log-likelihood of each option by the number of characters. In the symbol setting, the question and all answer options are provided together in a multiple-choice format. The model selects the answer by choosing the symbol (e.g., A, B, C, and D) with the highest log-likelihood. Since all symbols have the same length, we evaluate performance using the standard accuracy (\textit{acc}). 

We assess models in both zero-shot and few-shot settings (1, 2, 5, and 10 shots). For the few-shot evaluations, we sample demonstration examples with three random seeds (1234, 308, 1028) and report the average performance across them. 

\paragraph{Supervised fine-tuning.}
Using the training set from \bench, we perform supervised fine-tuning (SFT) on 35 models with fewer than 20 billion parameters. To prevent catastrophic forgetting~\citep{kirkpatrick2017overcoming}, where a model loses previously learned knowledge while acquiring new information, and to enable parameter-efficient training, we apply Low-Rank Adaptation (LoRA)~\citep{hulora}. Specific configurations can be found in Appendix~\ref{app:config}.

\begin{figure}[!t]
\centering
\begin{footnotesize}
\begin{minipage}{\linewidth}
\centering
\includegraphics[page=5,width=\linewidth]{figures/result_modelSize_fontup3.pdf}
\end{minipage}
\vspace{\baselineskip}

\begin{minipage}[t]{0.49\linewidth}
\centering
\includegraphics[page=1,width=\linewidth]{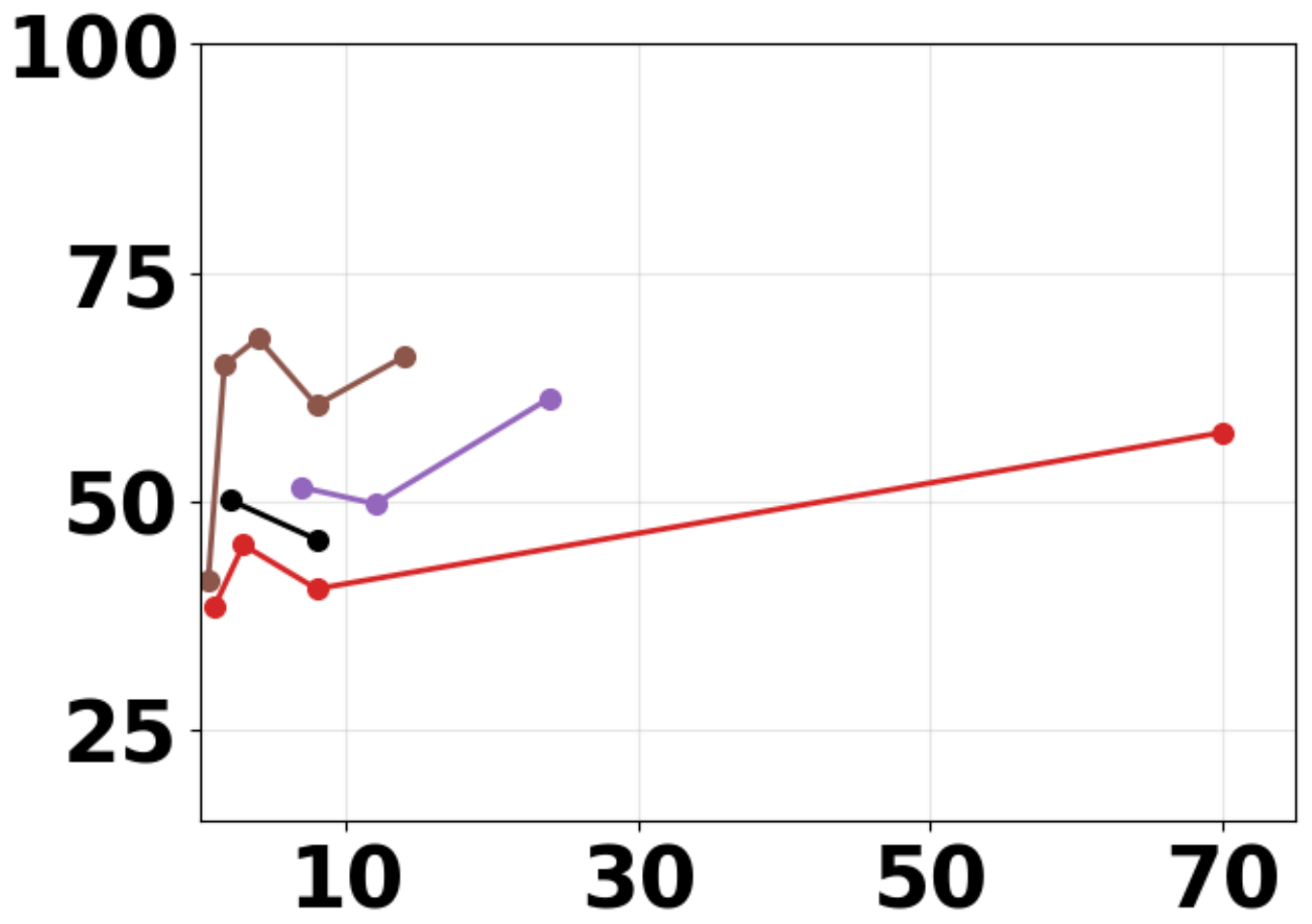}
(a) Base models (cloze).
\end{minipage}
\hfill
\begin{minipage}[t]{0.49\linewidth}
\centering
\includegraphics[page=2,width=\linewidth]{figures/result_modelSize_fontup3.pdf}
(b) Base models (symbol).
\end{minipage}
\vspace{\baselineskip}

\begin{minipage}[t]{0.49\linewidth}
\centering
\includegraphics[page=3,width=\linewidth]{figures/result_modelSize_fontup3.pdf}
(c) Instruction-tuned models (cloze).
\end{minipage}
\hfill
\begin{minipage}[t]{0.49\linewidth}
\centering
\includegraphics[page=4,width=\linewidth]{figures/result_modelSize_fontup3.pdf}
(d) Instruction-tuned models (symbol).
\end{minipage}
\end{footnotesize}
\caption{Model performance across different model sizes on zero-shot setting. The horizontal axis represents model size (in billions of parameters), and the vertical axis indicates performance (\textit{acc} or \textit{acc\_norm}).}
\label{fig:model_size}
\end{figure}

\subsection{Results}
We analyze the results by examining trends in model size and by comparing base models with instruction-tuned models. Additionally, we investigate the performance improvements achieved through SFT. Detailed evaluation results can be found in Appendix~\ref{app:eval_results}. 

\paragraph{Model size.} 
Our observations reveal a consistent trend across both Korean and non-Korean models: within each model family, larger models tend to achieve better performance. This pattern is evident in both evaluation formats—the cloze and symbol settings—indicating that increases in model size are generally associated with improved handling of negation (see Figure~\ref{fig:model_size}). This finding contrasts with earlier research~\citep{truong2023language}, which suggested that larger models are less sensitive to negation. In contrast, our results indicate that larger models are better equipped to move beyond superficial cues, making them more robust when addressing sentence-level negation.

It is important to note that this improvement is not strictly monotonic. We observe a noticeable slowdown or even a temporary decline in performance, especially among models with 8 to 12 billion parameters. This non-monotonic trend suggests that we are in a transitional phase of model scaling, where increasing capacity does not immediately lead to better handling of Korean negation. We hypothesize that this pattern indicates a mismatch between the emerging representational complexity and the level of linguistic supervision needed to master the nuanced aspects of negation.

\begin{figure}[!t]
\centering
\begin{footnotesize}
\begin{minipage}{\linewidth}
\centering
\includegraphics[page=3,width=\linewidth]{figures/result_instruct7.pdf}
\end{minipage}
\vspace{\baselineskip}
\begin{minipage}[t]{0.49\linewidth}
\centering
\includegraphics[page=1,width=\linewidth]{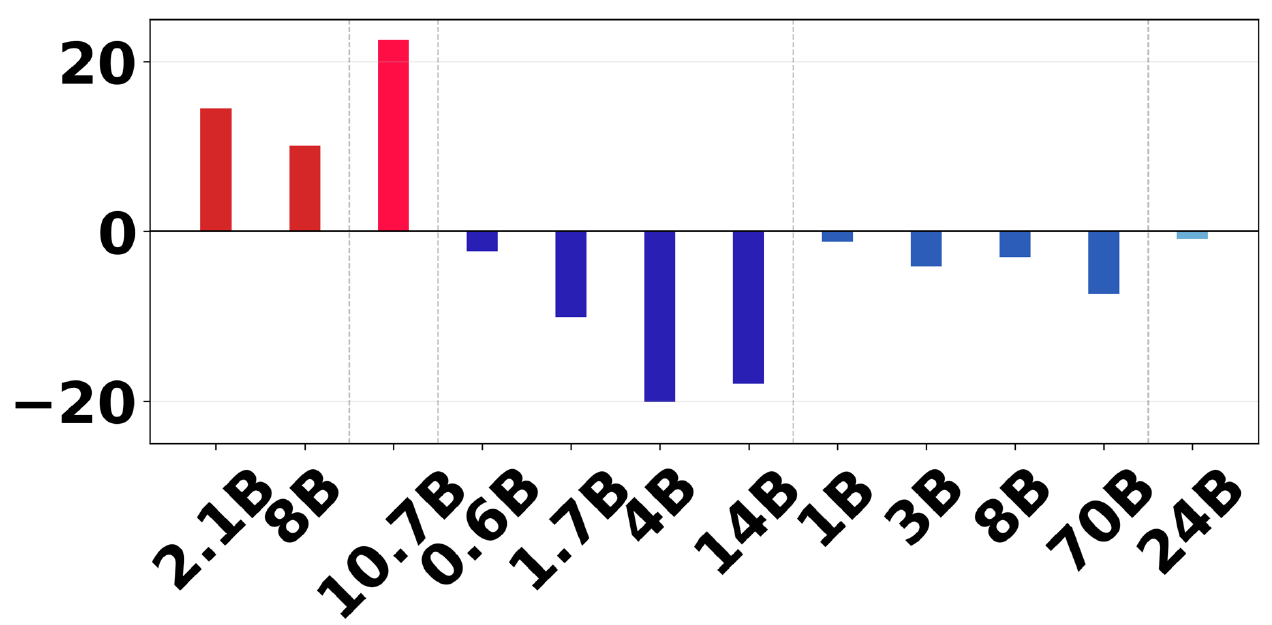}
(a) Cloze.
\end{minipage}
\hfill
\begin{minipage}[t]{0.49\linewidth}
\centering
\includegraphics[page=2,width=\linewidth]{figures/result_instruct7.pdf}
(b) Symbol.
\end{minipage}
\end{footnotesize}
\vspace{-1\baselineskip}
\caption{Performance change of instruction-tuned models relative to their base counterparts. The vertical axis represents the difference between the performance of instruction-tuned models and that of the corresponding base models.}
\label{fig:instruct_result}
\end{figure}

\paragraph{Instruction-tuned models.} 
In Figure~\ref{fig:instruct_result}, it is observed that instruction tuning improves general performance in the symbol setting. However, instruction tuning of non-Korean models often degrades performance in the cloze setting, suggesting that instruction tuning overemphasizes symbol-style MCQA formats. This bias improves format proficiency but reduces robustness to negation when evaluated on low-resource languages such as Korean. 

These findings suggest that instruction tuning may exacerbate the \textit{curse of multilinguality}~\citep{conneau-etal-2020-emerging,wang-etal-2020-negative}. Fine-tuning on high-resource languages can negatively affect performance on low-resource languages, and this effect is especially pronounced in understanding negation.

\paragraph{Error Analysis.}
\input{tables/5_2_error_analysis}

Table~\ref{tab:5_2_error_analysis} presents the distribution of incorrect choices selected by representative Korean and non-Korean models on \bench. The full error analysis results for all models are presented in Appendix~\ref{app:error_analysis}. In the cloze setting, more than 90\% of the errors are concentrated on local negation options. This pattern consistently holds across model families, instruction-tuning settings, model sizes, and overall performance levels, as well as across both Korean and non-Korean models.

In the symbol-based setting, the concentration on local negation is somewhat reduced compared to the cloze setting, but it still remains the dominant error type. These results suggest that models do not fully capture the semantic effect of negation in Korean; rather, they tend to focus on the surface-level application of negation markers.

\paragraph{Effect of SFT.} 
\input{tables/5_2_sft_brief}

SFT is conducted in two settings: cloze and symbol. We find that both approaches yield performance comparable to or even better than the original 10-shot results, without compromising performance on other tasks such as ARC~\citep{clark2018think}, HellaSwag~\citep{zellers2019hellaswag}, and Winogrande~\citep{sakaguchi2021winogrande}, as shown in Table~\ref{tab:sft_brief}. Additionally, our experiments reveal an apparent asymmetry in transferability between the two formats. Fine-tuning on the cloze format leads to an average improvement of 10.5\% in symbol-format evaluation, whereas fine-tuning on the symbol format only results in a 6.4\% enhancement in cloze-format evaluation. This asymmetry suggests that training models to generate negated sentences (required in the cloze format) provides a richer and more generalizable supervision signal than merely selecting from pre-generated options (e.g., A, B, C, and D). 

Cloze-style fine-tuning requires models to generate the correct standard negation of a sentence, which involves identifying the negation scope and applying appropriate Korean negation to the main predicate. In contrast, symbol-based formulations reduce the negation task to a label selection problem, providing a limited training signal for learning how negation is constructed. 


Fine-tuning on \bench also improves broader contextual understanding, as evidenced by substantial gains on KoBest BoolQ, which we previously examined in Section~\ref{sec:performance}. Consistent with our negation results, cloze-style fine-tuning yields larger improvements than symbol-based fine-tuning, highlighting the importance of generation-based supervision for learning negation.

%% file: tables/5_2_error_analysis.tex
\begin{table*}[htbp]
\centering
\footnotesize
\resizebox{\textwidth}{!}{%
\begin{tabular}{@{}llcc|ccc@{}}
\toprule
& \textbf{model name} & \makecell{\textbf{evaluation}\\\textbf{method}} & \textbf{performance} & \multicolumn{3}{c@{}}{\textbf{incorrect choice distribution}} \\
\cmidrule(l){5-7}
& & & & \textbf{local negation(\%)} & \textbf{contradiction(\%)} & \textbf{paraphrase(\%)} \\
\midrule

\multirow{8}{*}{\textbf{Non-Korean}}
& \multirow{2}{*}{\textbf{Qwen3-8B-Base}} 
& \textbf{cloze} & 56.31 & 94.83 & 3.74 & 1.43 \\
& & \textbf{symbol} & 85.44 & 85.56 & 8.02 & 6.42 \\
\addlinespace[2pt]


& \multirow{2}{*}{\textbf{Llama-3.1-8B}} 
& \textbf{cloze} & 39.56 & 93.17 & 4.51 & 2.32 \\
& & \textbf{symbol} & 53.66 & 53.28 & 12.27 & 34.45 \\
\addlinespace[2pt]

& \multirow{2}{*}{\textbf{Mistral-7B-Instruct-v0.3}} 
& \textbf{cloze} & 63.16 & 93.66 & 4.65 & 1.69 \\
& & \textbf{symbol} & 72.90 & 75.86 & 6.03 & 18.10 \\
\addlinespace[2pt]

& \textbf{gpt-4.1} 
& \textbf{symbol} & 92.13 & 97.03 & 1.98 & 0.99 \\
\addlinespace[2pt]

& \textbf{claude-sonnet-4-5-20250929} 
& \textbf{symbol} & 98.05 & 72.00 & 28.00 & 0.00 \\
\midrule

\multirow{10}{*}{\textbf{Korean}}
& \multirow{2}{*}{\textbf{Midm-2.0-Base-Instruct (11.5B)}} 
& \textbf{cloze} & 61.21 & 94.38 & 4.42 & 1.20 \\
& & \textbf{symbol} & 59.50 & 50.77 & 30.77 & 18.46 \\
\addlinespace[2pt]


& \multirow{2}{*}{\textbf{kanana-1.5-8b-instruct-2505}} 
& \textbf{cloze} & 53.50 & 94.64 & 4.19 & 1.17 \\
& & \textbf{symbol} & 56.70 & 51.44 & 13.67 & 34.89 \\
\addlinespace[2pt]

& \multirow{2}{*}{\textbf{SOLAR-10.7B-v1.0}} 
& \textbf{cloze} & 40.50 & 93.98 & 4.71 & 1.31 \\
& & \textbf{symbol} & 41.98 & 44.83 & 14.23 & 40.94 \\
\addlinespace[2pt]


& \multirow{2}{*}{\textbf{EXAONE-4.0-32B}} 
& \textbf{cloze} & 50.16 & 96.41 & 3.12 & 0.47 \\
& & \textbf{symbol} & 81.46 & 90.76 & 5.88 & 3.36 \\

& \multirow{2}{*}{\textbf{A.X-4.0 (72B)}} 
& \textbf{cloze} & 73.83 & 91.67 & 5.65 & 2.68 \\
& & \textbf{symbol} & 95.72 & 87.27 & 7.27 & 5.45 \\

\bottomrule
\end{tabular}%
}
\caption{Incorrect choice distributions of Korean and non-Korean models under zero-shot evaluation.}
\label{tab:5_2_error_analysis}
\end{table*}

%% file: tables/5_2_sft_brief.tex
\begin{table*}[!t]
\centering
\resizebox{\textwidth}{!}{%
\setlength{\tabcolsep}{2pt} 
\begin{tabular}{@{}c|c|c|c|c|c|c|c|c|c|c@{}}
\toprule
 & \multicolumn{2}{c|}{\textbf{\bench}}
 & \multicolumn{2}{c|}{\textbf{KMMLU}}
 & \multicolumn{2}{c|}{\textbf{BoolQ}}
 & \multicolumn{2}{c|}{\textbf{ARC}}
 & \multirow{2}{*}{\textbf{Hellaswag}}
 & \multirow{2}{*}{\textbf{Winogrande}} \\
\cmidrule(lr){2-3}
\cmidrule(lr){4-5}
\cmidrule(lr){6-7}
\cmidrule(lr){8-9}
 & \textbf{Cloze}
 & \textbf{Symbol}
 & \textbf{Negative}
 & \textbf{Affirmative}
 & \textbf{Original}
 & \textbf{Negated}
 & \textbf{Easy}
 & \textbf{Challenge}
 &  &  \\
\midrule
\begin{tabular}{c}
Cloze-style \\ fine-tuning 
\end{tabular} & 85.1 (+34.2) & 69.6 (+10.5) & 61.0 (+0.4) & 62.3 (-0.1) & 66.7 (+3.0) & 53.4 (+0.9) & 71.5 (+0.5) & 49.2 (+0.2) & 70.7 (+0.1) & 67.4 (+0.1) \\
\begin{tabular}{c}
Symbol-style \\ fine-tuning 
\end{tabular} & 57.3 (+6.4) & 89.8 (+30.7) & 61.2 (+0.7) & 62.5 (+0.1) & 65.5 (+1.8) & 52.8 (+0.3) & 71.4 (+0.4) & 49.2 (+0.2) & 70.7 (+0.1) & 67.3 (+0.0) \\
\bottomrule
\end{tabular}%
}
\vspace{-0.5\baselineskip}
\caption{Average model performance after SFT on \bench and other tasks. Values in parentheses indicate the performance gain relative to the baseline.}
\label{tab:sft_brief}
\end{table*}

%% file: latex/6_conclusion.tex
\section{Conclusion}

In this paper, we demonstrate that LLMs struggle with negation in Korean and introduce \bench, a benchmark for evaluating sentence-level negation understanding. Inspired by Thunder-NUBench, \bench covers a broad range of Korean-specific negation types and structures and closely matches their empirical distribution. Evaluating 47 LLMs, we find that larger models are generally more robust, while multilingual instruction tuning can degrade negation performance in low-resource languages like Korean. Error analysis indicates that most errors concentrate on local negation, suggesting reliance on surface-level negation cues rather than true semantic understanding. Fine-tuning experiments further show that a cloze-based format provides more effective supervision than symbol-based alternatives for improving negation understanding. \bench is publicly available at \url{https://champ.snu.ac.kr/}.

%% file: latex/limitations.tex
\section*{Limitations}

Lexical negation is inherently relational, as forms such as "있다" (exist) and "없다" (not exist) constitute negation only with respect to each other, rather than in isolation. As a result, in our corpus-level statistical analysis, we primarily focused on syntactic negation markers and did not identify or count lexical negation. However, \bench itself includes a number of instances involving lexical negation, as complementary antonyms are permitted when constructing standard negation. Both syntactic and lexical negation are therefore represented in \bench.

Furthermore, \bench does not simply require models to interpret naturally occurring negated sentences; instead, it asks them to identify the option that correctly corresponds to the negation of a given original sentence. This design is motivated by our goal of capturing a fundamental aspect of negation—the operation that maps a proposition $P$ to its counterpart $\neg P$—in \bench. As a result, our benchmark does not evaluate negation understanding in more naturalistic settings, where negation appears in natural contexts. In particular, it does not assess how models comprehend and reason over negation in context. Developing more naturalistic evaluation settings that examine how models understand negation in context remains an important direction for future work.

%% file: latex/ethics.tex
\section*{Ethical Considerations}
This work does not rely on crowdsourcing. Instead, all data included in \bench were manually reviewed by the authors to ensure high quality, relevance, and compliance with ethical standards. All datasets and tools used for training and evaluation are publicly available and were utilized in accordance with their respective licenses.

When using OpenAI’s text generation models, we exercise additional caution to prevent the inclusion of harmful, biased, or privacy-violating content. All generated examples undergo manual inspection to ensure they meet ethical and safety requirements. In particular, we verify that the final dataset contains no personally identifiable information or offensive material.

The human evaluation was approved by the Institutional Review Board (IRB No. 2512/004-012). All participants received sufficient information about the study and were given adequate rest time, ensuring that the evaluation was conducted in an ethically responsible manner.

The \bench dataset is made available under the CC BY-NC-SA 4.0 license to support transparent and reproducible research and to facilitate responsible reuse. We believe that our work provides a meaningful step toward building more trustworthy and interpretable language models.

%% file: latex/acknowlegments.tex
\section*{Acknowlegments}
This work was partially supported by the National Research Foundation of Korea (NRF) under Grant No. RS-2023-00222663 (Center for Optimizing Hyperscale AI Models and Platforms), and by the Institute for Information and Communications Technology Promotion (IITP) under Grant No. 2018-0-00581 (CUDA Programming Environment for FPGA Clusters) and No. RS-2025-02304554 (Efficient and Scalable Framework for AI Heterogeneous Cluster Systems), all funded by the Ministry of Science and ICT (MSIT) of Korea. It was also partially supported by the Korea Health Industry Development Institute (KHIDI) under Grant No. RS-2025-25454559 (Frailty Risk Assessment and Intervention Leveraging Multimodal Intelligence for Networked Deployment in Community Care), funded by the Ministry of Health and Welfare (MOHW) of Korea. Additional support was provided by the BK21 Plus Program for Innovative Data Science Talent Education (Department of Data Science, Seoul National University, No. 5199990914569) and the BK21 FOUR Program for Intelligent Computing (Department of Computer Science and Engineering, Seoul National University, No. 4199990214639), both funded by the Ministry of Education (MOE) of Korea. This work was also partially supported by the Artificial Intelligence Industrial Convergence Cluster Development Project, funded by the MSIT and Gwangju Metropolitan City. Research facilities were provided by the Institute of Computer Technology (ICT) at Seoul National University.

%% file: latex/appendix.tex
\appendix

\section{Negative Expressions in Korean}
\label{app:neg_korean}

\subsection{Syntactic Negation}
\input{tables/app_negation_in_korean}
There are three main types of syntactic negation in Korean, each characterized by distinct meanings and syntactic constraints(Table~\ref{tab:neg_type} shows details and examples). In constructing \bench, we carefully adhered to these constraints and incorporated all three types.

\paragraph{안 계열 (\textit{An} type).} The \textit{An} type is used to either simply reverse the truth value or to negate the subject's volition. It cannot be used in sentences whose predicates involve actions that presuppose the subject's ability (e.g., "\textit{알다}"(know), "\textit{견디다}"(endure)). The \textit{An} type is divided into short-form negation, where a negative adverb "\textit{안}" precedes the predicate, and long-form negation, where a negative auxiliary verb "\textit{-지 않-}" follows the predicate. 

\paragraph{못 계열 (\textit{Mot} Type).} \textit{Mot} type expresses the subject's inability to perform an action (corresponding to \textit{cannot} in English). It cannot be used with predicates that are adjectival inflections(e.g., "\textit{착하다}"(kind), "\textit{아름답다}"(beautiful)), as the notion of inability cannot be semantically ascribed to stative predicates. Like the \textit{An} type, it has both a short form, where "\textit{못}" precedes the predicate, and a long form, where the negative auxiliary verb "\textit{-지 못하-}" follows the predicate.

\paragraph{말다 (\textit{Malda}).} In imperative or hortative sentence, prohibitive negation is realized by attaching the auxiliary verb "\textit{-지 말-}" after the predicate.

\paragraph{Constraints of short-form negation.}
Short-form negation realized by negative adverbs cannot generally be applied to predicates that are compounds or derived forms. For example, expressions such as “안 아름답다” and “못 공부했다” are considered awkward, whereas their long-form counterparts “아름답지 않다” and “공부하지 못했다” are natural.
However, this restriction is not absolute. When compound predicates are formed through auxiliary connective endings such as -아/어-, short-form negation is permitted (e.g., “안 들어가다”). In addition, short-form negation is also allowed when the predicate is a derived form created by passive suffixes (-이-, -히-, -리-, -기-) or causative suffixes (-이-, -히-, -리-, -기-, -우-, -구-, -추-).

\subsection{Lexical Negation}
Lexical negation in Korean is realized either by attaching a negative prefix or by using a complementary antonym. Negative prefixes (e.g., "비-", "불-", "미-", "무-", "몰-") are attached to nominals. For example, the noun "공평" (fairness) combines with the negative prefix "불-" to form "불공평" (unfairness). In addition, certain derivational suffixes (e.g., "-하다") attach to nouns, ideophones, or mimetic words to form predicates. For instance, "공평" (fairness) becomes "공평하다" (is fair), which functions as a predicate in a sentence. The same process also applies to "불공평" (unfairness), yielding "불공평하다" (is unfair), which likewise serves as a predicate.

Since \bench focuses on negation at the predicate level, we allow prefix-based negation only when a predicate is formed as a "nominal + 하다" (e.g., "공평하다"(is fair) $\rightarrow$ "{\color{red}불}공평하다"(is unfair)) construction.

Antonyms can be broadly categorized into three types: 
\begin{itemize}
    \item{complementary antonyms: mutually exclusive pairs with no intermediate states (e.g., 살다(alive) / 죽다(dead))}
    \item{gradable antonyms: pairs of words that denote opposite ends of a continuous scale, allowing for intermediate degrees between the two extremes (e.g., 덥다(hot) / 춥다(cold))}.
    \item{relational antonyms: pairs of words that describe a reciprocal
relationship where one implies the existence of the other (e.g., 사다(buy) / 팔다(sell)).}
\end{itemize}
However, from the perspective of negation that completely reverses the meaning of the original sentence, only complementary antonyms constitute true negation. Accordingly, \bench permits only complementary antonyms when applying lexical negation to predicates.

\section{Sentence Structure in Korean}
\label{app:korean_sentence}

\input{tables/app_sentence_korean}
In both corpus analysis and the construction of \bench, we carefully adhere to the characteristics of Korean sentence structure. In this section, we describe the sentence structure of Korean. 

Korean sentence structure can be broadly divided into 홑문장(simple sentences) and 겹문장(complex sentences). A simple sentence contains a single main(i.e., independent) clause, whereas a complex sentence consists of two or more clauses, featuring multiple subject–predicate structures.
Complex sentences are further classified into 이어진 문장(connected sentences) and 안긴 문장(sentences with embedded clause), depending on how the clauses are combined. Table~\ref{tab:sentence_structure} presents the detailed descriptions and examples. 

\subsection{Connected Sentences}
A connected sentence is formed by combining two or more clauses through connective endings(e.g., -고, -거나, -(어/아)서).
Coordinated sentences can be further divided into:
\begin{itemize}
    \item{대등하게 이어진 문장(Coordinated sentences), where two or more main clauses are linked in an equal relationship,}
    \item{종속적으로 이어진 문장(sentences with subordinate clauses), where a subordinate clause is combined with a main clause in a dependent relationship.}
\end{itemize}

\subsection{Sentences with Embedded Clause}
In Korean, a clause can be embedded within another matrix clause, functioning as a single grammatical constituent. In this paper, we treat the matrix clause as the main clause and the embedded clause as the dependent clause. This distinction is motivated by the design principle of \bench: clauses are categorized as main clauses if they can function as a complete sentence on their own, and as dependent clauses otherwise.

Embedded sentences are classified according to the grammatical role of the embedded clause, including:
\begin{itemize}
    \item{명사절을 안은 문장(sentences with embedded noun clauses)}
    \item{관형절을 안은 문장(sentences with embedded adnominal clauses)}
    \item{인용절을 안은 문장(sentences with embedded quotation clauses)}
    \item{부사절을 안은 문장(sentences with embedded adverbial clauses)}
    \item{서술절을 안은 문장(sentences with embedded predicative clauses)}.
\end{itemize}

\section{Details of the corpus}
\label{app:corpus}

\input{tables/app_corpus}

This section describes the corpus used for analyzing the statistical distribution of negation in Korean, as introduced in Section~\ref{subsec:3_2_dists}. We use the dataset titled ``한국어 성능이 개선된 초거대 AI 언어모델 개발 및 데이터'' (Dataset and Large Language Models with Improved Korean Performance). It consists of approximately 2.28 billion whitespace-delimited Korean word units(어절; Eojel) and is publicly released by The Open AI Dataset Project\citep{AiHub}.

The corpus contains both spoken and written-style texts and spans a wide range of topics and domains. Table~\ref{tab:corpus_sentence_type} presents the distribution of sentence types in the corpus, and Table~\ref{tab:corpus_category} shows the topic-wise distribution of sentences. The diversity of the corpus ensures that the observed statistical patterns of negation reflect real-world usage of Korean negation phenomena.

\section{Details of Constructing \bench}
\label{app:construct}
This section describes the data construction procedure for \bench, including sentence-pair extraction, sentence merging, and the generation of contradiction and paraphrase options. The prompts used in this process were originally written in Korean, and English translations are provided in the corresponding figures.

\paragraph{Splitting original text into sentence pairs.}
We first crawled the entire Korean Wikipedia and segmented the raw text into individual sentences using the \texttt{split\_sentences} function from the \texttt{KSS} package, a rule-based Korean sentence splitter. From the crawled corpus, we extracted approximately 4 million consecutive sentence pairs. Since our target benchmark size was about 5,000 instances, we did not use all extracted pairs. Instead, we randomly sampled 5,000 sentence pairs from this pool.

\begin{figure*}[!htbp]
\centering
\includegraphics[width=\linewidth]{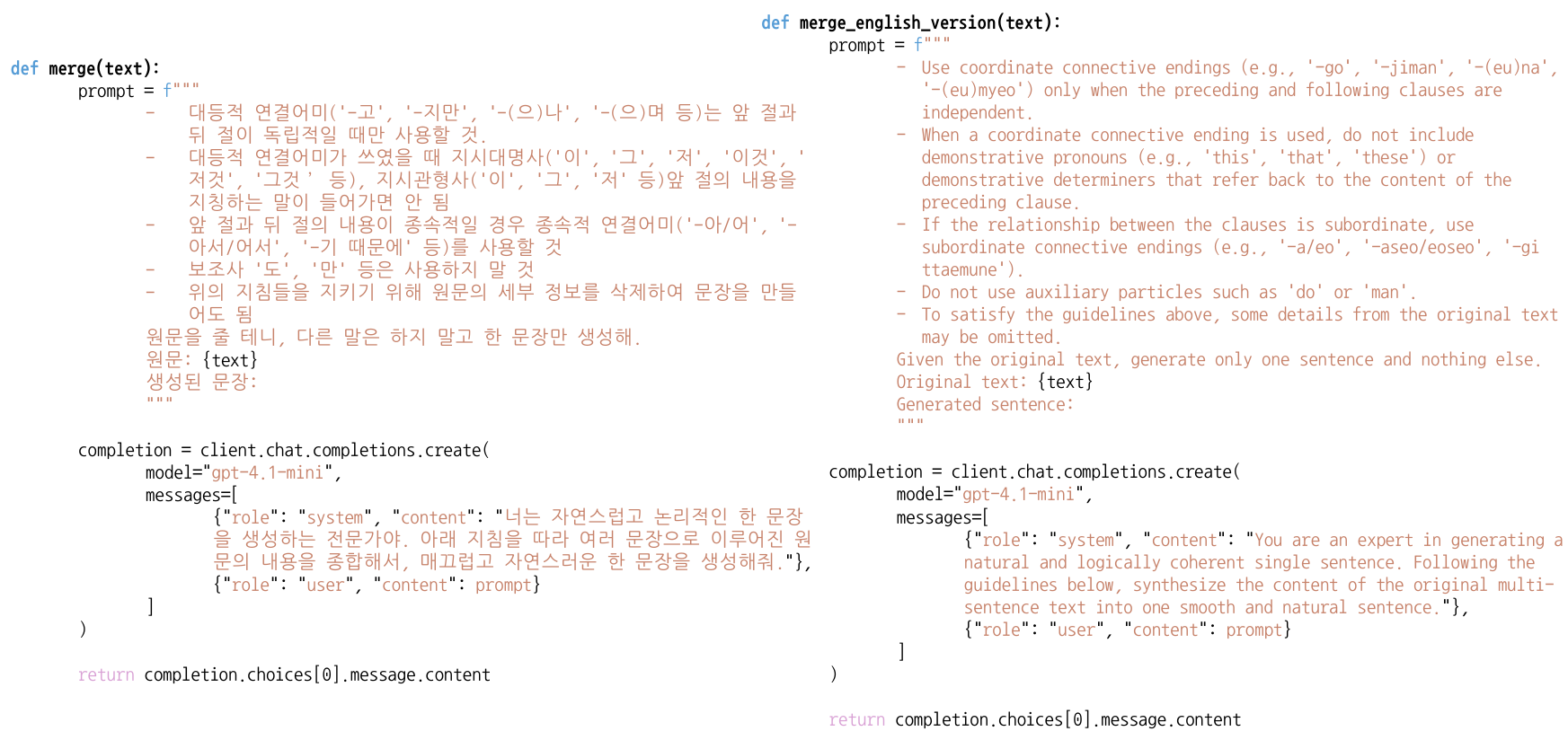}
\caption{Korean prompt used to merge a pair of sentences into a single natural and well-formed sentence, with an English translation.}
\label{fig:code_merge}
\end{figure*}

\paragraph{Merging sentence pairs into a single sentence.}
Each two-sentence unit was converted into a single, natural, and well-formed sentence using the GPT-4.1 mini model with OpenAI API. This step was designed to avoid overly short or trivial source sentences and to obtain contexts rich enough for constructing benchmark instances. During generation, we instructed the model to preserve the original meaning while producing a grammatically coherent sentence and avoiding unnecessary discourse markers or unnatural referential expressions. The Korean and English versions of the prompt used for this step are shown in Figure~\ref{fig:code_merge}. The generated outputs were later manually checked and corrected when needed.

\begin{figure*}[!htbp]
\centering
\includegraphics[width=\linewidth]{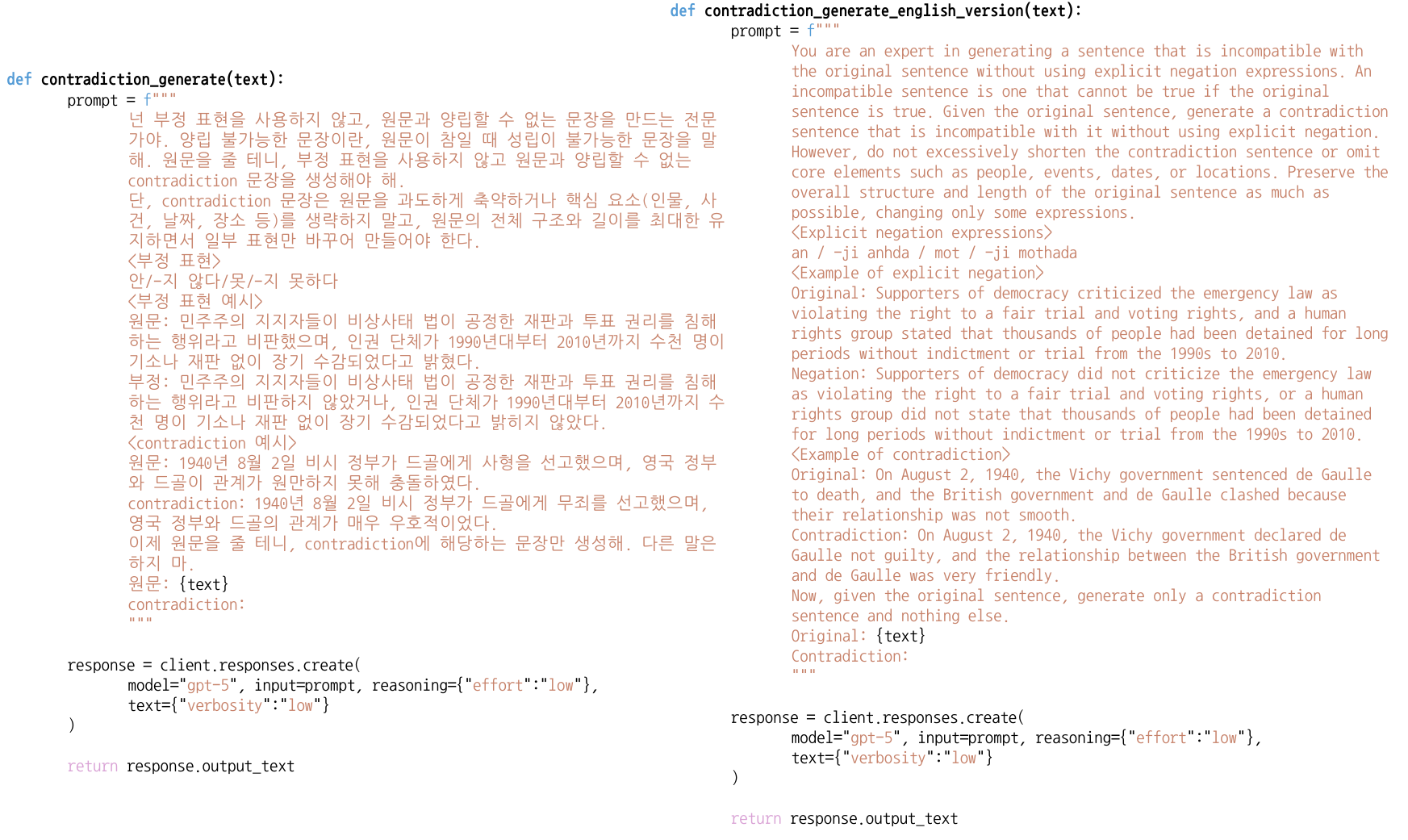}
\caption{Korean prompt used to generate a contradiction sentence that is incompatible with the original sentence without explicit negation, with an English translation.}
\label{fig:code_contradiction}
\end{figure*}

\begin{figure*}[!htbp]
\centering
\includegraphics[width=\linewidth]{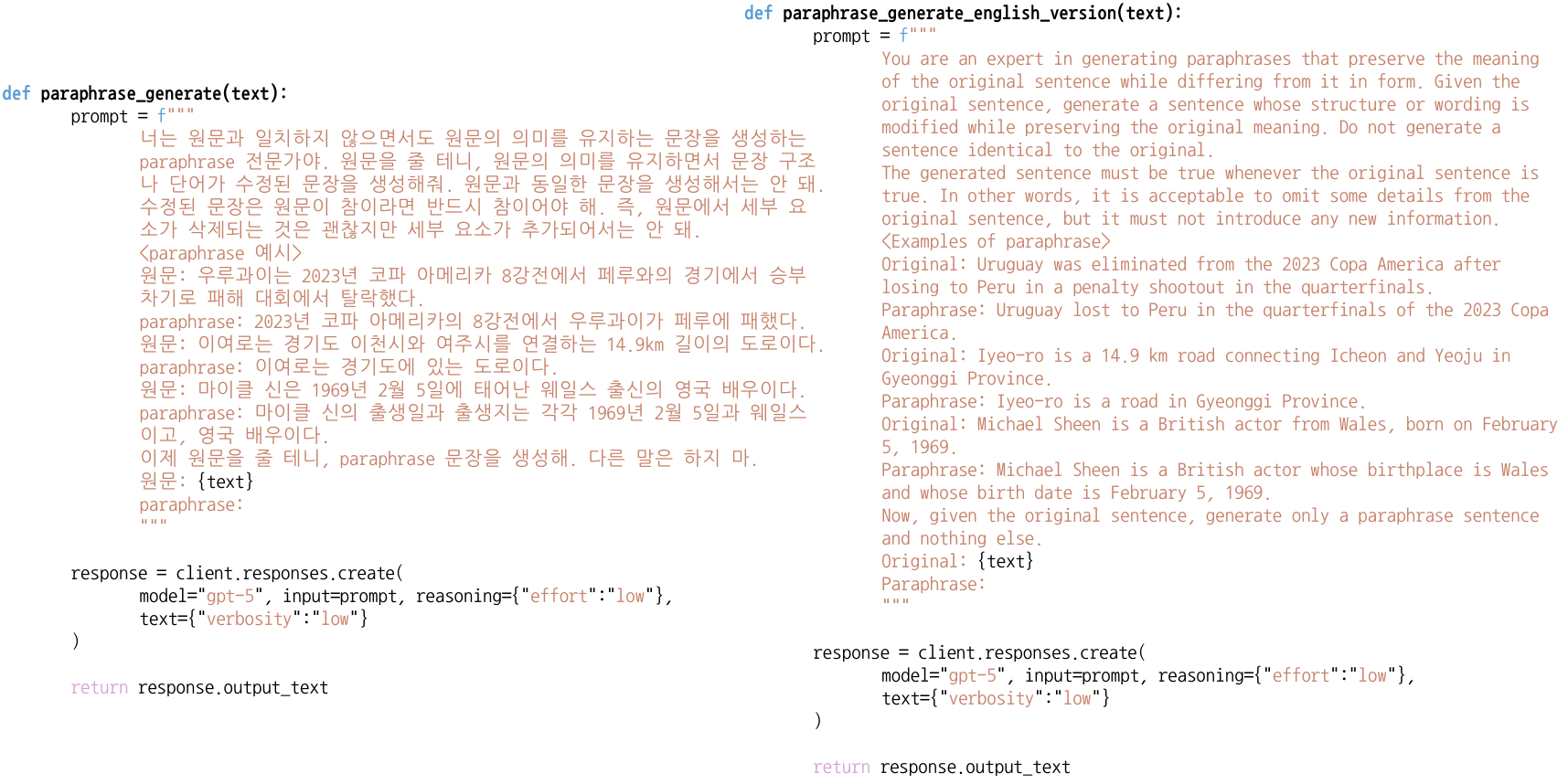}
\caption{Korean prompt used to generate a paraphrase that preserves the original meaning while changing the surface form, with an English translation.}
\label{fig:code_paraphrase}
\end{figure*}

\paragraph{Generating contradiction and paraphrase options.}
We generated contradiction and paraphrase candidates from the merged original sentences using the GPT-5 model with OpenAI API. For contradiction, we instructed the model to produce a sentence incompatible with the original one without using explicit negation markers, while preserving the overall structure as much as possible. The corresponding Korean and English prompts are shown in Figure~\ref{fig:code_contradiction}. For paraphrase, we instructed the model to preserve the original meaning while changing the surface form without adding new information. The corresponding prompts are shown in Figure~\ref{fig:code_paraphrase}. All generated candidates were manually reviewed and refined by the authors.

\section{Details of the Review Process}
\label{app:review}
\bench is constructed through a thorough review process among the authors. In this section, we provide the protocol in reviewing and how we deal with inter-annotators' disagreements.

\subsection{Human Review Protocol}
\bench is constructed through a thorough review process conducted by the authors. 
In this section, we describe the review protocol and how inter-annotator disagreements are identified and resolved.
\paragraph{Independent task allocation and review.}
Each author is assigned a distinct subset of the dataset. For each instance, the author first finalizes the original sentence and then generates its corresponding standard negation and local negation. In addition, contradiction and paraphrase options initially generated by OpenAI API\citep{OpenAI_2025} are manually reviewed and revised to ensure linguistic correctness and semantic validity.

After an author completes their assigned subset, all instances undergo independent cross-checking by another author. The reviewer verifies whether each option is appropriate with respect to the original sentence and records all cases of disagreement.

\paragraph{Consensus-building and revision.} All disputed instances are discussed in weekly regular meetings attended by all authors. During these meetings, we jointly assess the validity of each disagreement and determine appropriate revisions when necessary. Through this consensus-based revision process, all issues are resolved and the final versions of the data are confirmed.

\subsection{Inter-annotator Agreement}
\input{tables/app_inter_annotation}
We conduct two rounds of cross-checking for all items, each performed by an independent reviewer.
Table~\ref{tab:inter_annotation} reports the proportion of items for which at least one label (original sentence, standard negation, local negation, contradiction, or paraphrase) was marked as a disagreement during each review stage.
All disagreement cases are resolved through consensus-building discussions in regular meetings, yielding a fully consistent final dataset.

\section{Details of the Models}
\label{app:models}
This section details the models evaluated in this paper and their implementation methods. For all evaluation, we use a 16-bit (bf16) quantized model.

\subsection*{Korean Models}
\begin{itemize}
\small
\item{Midm-2.0}
\begin{itemize}[leftmargin=6mm,itemsep=0pt,parsep=0pt,topsep=0pt]
    \item{\href{https://huggingface.co/K-intelligence/Midm-2.0-Mini-Instruct}{Midm-2.0-Mini-Instruct} (2.3B)}
    \item{\href{https://huggingface.co/K-intelligence/Midm-2.0-Base-Instruct}{Midm-2.0-Base-Instruct} (11.5B)}
\end{itemize}
\item{EXAONE}
\begin{itemize}[leftmargin=6mm,itemsep=0pt,parsep=0pt,topsep=0pt]
    \item{\href{https://huggingface.co/LGAI-EXAONE/EXAONE-4.0-1.2B}{EXAONE-4.0-1.2B}}
    \item{\href{https://huggingface.co/LGAI-EXAONE/EXAONE-4.0-32B}{EXAONE-4.0-32B}}
    \item{\href{https://huggingface.co/LGAI-EXAONE/EXAONE-Deep-2.4B}{EXAONE-Deep-2.4B}}
    \item{\href{https://huggingface.co/LGAI-EXAONE/EXAONE-Deep-7.8B}{EXAONE-Deep-7.8B}}
    \item{\href{https://huggingface.co/LGAI-EXAONE/EXAONE-Deep-32B}{EXAONE-Deep-32B}}
\end{itemize}
\item{kanana-1.5}
\begin{itemize}[leftmargin=6mm,itemsep=0pt,parsep=0pt,topsep=0pt]
    \item{\href{https://huggingface.co/kakaocorp/kanana-1.5-2.1b-base}{kanana-1.5-2.1b-base}}
    \item{\href{https://huggingface.co/kakaocorp/kanana-1.5-2.1b-instruct-2505}{kanana-1.5-2.1b-instruct-2505}}
    \item{\href{https://huggingface.co/kakaocorp/kanana-1.5-8b-base}{kanana-1.5-8b-base}}
    \item{\href{https://huggingface.co/kakaocorp/kanana-1.5-8b-instruct-2505}{kanana-1.5-8b-instruct-2505}}
\end{itemize}
\item{HyperCLOVAX-SEED}
\begin{itemize}[leftmargin=6mm,itemsep=0pt,parsep=0pt,topsep=0pt]
    \item{\href{https://huggingface.co/naver-hyperclovax/HyperCLOVAX-SEED-Text-Instruct-0.5B}{HyperCLOVAX-SEED-Text-Instruct-0.5B}}
    \item{\href{https://huggingface.co/naver-hyperclovax/HyperCLOVAX-SEED-Text-Instruct-1.5B}{HyperCLOVAX-SEED-Text-Instruct-1.5B}}
    \item{\href{https://huggingface.co/naver-hyperclovax/HyperCLOVAX-SEED-Think-14B}{HyperCLOVAX-SEED-Think-14B}}
\end{itemize}
\item{A.X-4.0}
\begin{itemize}[leftmargin=6mm,itemsep=0pt,parsep=0pt,topsep=0pt]
    \item{\href{https://huggingface.co/skt/A.X-4.0-Light}{A.X-4.0-Light} (7.2B)}
    \item{\href{https://huggingface.co/skt/A.X-4.0}{A.X-4.0} (72B)}
\end{itemize}
\item{SOLAR}
\begin{itemize}[leftmargin=6mm,itemsep=0pt,parsep=0pt,topsep=0pt]
    \item{\href{https://huggingface.co/upstage/SOLAR-10.7B-v1.0}{SOLAR-10.7B-v1.0}}
    \item{\href{https://huggingface.co/upstage/SOLAR-10.7B-Instruct-v1.0}{SOLAR-10.7B-Instruct-v1.0}}
\end{itemize}
\end{itemize}

\subsection*{Non-Korean Models}
\begin{itemize}
\small
\item{Qwen3}
\begin{itemize}[leftmargin=6mm,itemsep=0pt,parsep=0pt,topsep=0pt]
    \item{\href{https://huggingface.co/Qwen/Qwen3-0.6B-Base}{Qwen3-0.6B-Base}}
    \item{\href{https://huggingface.co/Qwen/Qwen3-0.6B}{Qwen3-0.6B}}
    \item{\href{https://huggingface.co/Qwen/Qwen3-1.7B-Base}{Qwen3-1.7B-Base}}
    \item{\href{https://huggingface.co/Qwen/Qwen3-1.7B}{Qwen3-1.7B}}
    \item{\href{https://huggingface.co/Qwen/Qwen3-4B-Base}{Qwen3-4B-Base}}
    \item{\href{https://huggingface.co/Qwen/Qwen3-4B}{Qwen3-4B}}
    \item{\href{https://huggingface.co/Qwen/Qwen3-8B-Base}{Qwen3-8B-Base}}
    \item{\href{https://huggingface.co/Qwen/Qwen3-8B}{Qwen3-8B}}
    \item{\href{https://huggingface.co/Qwen/Qwen3-14B-Base}{Qwen3-14B-Base}}
    \item{\href{https://huggingface.co/Qwen/Qwen3-14B}{Qwen3-14B}}
    \item{\href{https://huggingface.co/Qwen/Qwen3-32B}{Qwen3-32B}}
\end{itemize}

\item{Llama 3.1 and Llama 3.2}
\begin{itemize}[leftmargin=6mm,itemsep=0pt,parsep=0pt,topsep=0pt]
    \item{\href{https://huggingface.co/meta-llama/Llama-3.1-8B}{Llama-3.1-8B}}
    \item{\href{https://huggingface.co/meta-llama/Llama-3.1-8B-Instruct}{Llama-3.1-8B-instruct}}
    \item{\href{https://huggingface.co/meta-llama/Llama-3.1-70B}{Llama-3.1-70B}}
    \item{\href{https://huggingface.co/meta-llama/Llama-3.1-70B-Instruct}{Llama-3.1-70B-instruct}}
    \item{\href{https://huggingface.co/meta-llama/Llama-3.2-1B}{Llama-3.2-1B}}
    \item{\href{https://huggingface.co/meta-llama/Llama-3.2-1B-Instruct}{Llama-3.2-1B-instruct}}
    \item{\href{https://huggingface.co/meta-llama/Llama-3.2-3B}{Llama-3.2-3B}}
    \item{\href{https://huggingface.co/meta-llama/Llama-3.2-3B-Instruct}{Llama-3.2-3B-instruct}}
\end{itemize}

\item{Mistral}
\begin{itemize}[leftmargin=6mm,itemsep=0pt,parsep=0pt,topsep=0pt]
    \item{\href{https://huggingface.co/mistralai/Mistral-7B-v0.3}{Mistral-7B-v0.3}}
    \item{\href{https://huggingface.co/mistralai/Mistral-7B-Instruct-v0.3}{Mistral-7B-Instruct-v0.3}}
    \item{\href{https://huggingface.co/mistralai/Mistral-Nemo-Base-2407}{Mistral-Nemo-Base-2407} (12B)}
    \item{\href{https://huggingface.co/mistralai/Mistral-Nemo-Instruct-2407}{Mistral-Nemo-Instruct-2407} (12B)}
    \item{\href{https://huggingface.co/mistralai/Mistral-Small-24B-Base-2501}{Mistral-Small-24B-Base-2501}}
    \item{\href{https://huggingface.co/mistralai/Mistral-Small-24B-Instruct-2501}{Mistral-Small-24B-Instruct-2501}}
\end{itemize}
\end{itemize}

\section{Configurations for Fine-Tuning}
\label{app:config}
We apply LoRA\citep{lora} with a rank of 8, targeting all linear layers. The scaling factor alpha is set to 32, and the dropout rate to 0.05. We use a batch size of 128 with gradient accumulation, cosine learning-rate scheduling, and bfloat16 precision. All training is conducted on 8 AMD MI250 GPUs, each with 64 GB of memory.

\section{Evaluation Results}
\label{app:eval_results}

\input{tables/app_kmmlu_boolq}

\input{tables/app_baseline_result}

\input{tables/app_sft_cloze_result}

\input{tables/app_sft_symbol_result}
This section presents our comprehensive evaluation results. We report accuracy(\textit{acc}) for KMMLU\citep{son2025kmmlu} under the symbol-based evaluation, \textit{acc} for BoolQ\citep{jang-etal-2022-kobest} and Winogrande\citep{sakaguchi2021winogrande} under the cloze-based evaluation, and length-normalized accuracy (\textit{acc\_norm}) for ARC\citep{clark2018think} and HellaSwag\citep{zellers2019hellaswag} under the cloze-based evaluation, following the default settings of the LM Evaluation Harness\citep{sutawika2025eleutherai}.
For API-based models such as GPT and Claude, we report only symbol-based performance measured by \texttt{exact\_match}, since these models provide only final textual outputs and do not expose token-level likelihoods. All evaluations is conducted on 8 AMD MI250 GPUs, each with 64 GB of memory.

Table~\ref{tab:kmmlu_boolq} presents the evaluation results on KMMLU using the methodology introduced in Section~\ref{sec:performance}, comparing questions containing negation with their affirmative counterparts. In addition, it reports performance on KoBest BoolQ and the corresponding results obtained when the questions are transformed into their negated forms.
Table~\ref{tab:baseline_result} reports the zero-shot and few-shot performance on \bench.
Tables~\ref{tab:sft_cloze1} and~\ref{tab:sft_cloze2} present model performance after fine-tuning with a focus on the cloze format.
Tables~\ref{tab:sft_symbol1} and~\ref{tab:sft_symbol2} report model performance across diverse tasks after fine-tuning targeting the symbol format.

\section{Error Analysis}
\label{app:error_analysis}
\input{tables/app_error_analysis}
This section presents detailed error analysis results. 
Table~\ref{tab:error_analysis_qwen_llama} reports the results for the Qwen3~\citep{yang2025qwen3} and Llama~\citep{grattafiori2024llama3herdmodels} model families. 
Table~\ref{tab:error_analysis_mistral_gpt_claude} presents the results for the Mistral~\citep{jiang2023mistral7b}, GPT-4.1~\citep{achiam2023gpt}, and Claude-Opus 4.5~\citep{ANTHROPIC_2025} families. 
Table~\ref{tab:error_analysis_korean} provides the error analysis results for all Korean models.

%% file: tables/app_negation_in_korean.tex
\begin{table*}[htbp]
\resizebox{\textwidth}{!}{%
\begin{tabular}{@{}cccc@{}}
\toprule
\multicolumn{2}{c}{\textbf{Negation Type}}                                 & \textbf{Affirmative Sentence}                                                                 & \textbf{Negative Sentence}                                                                                 \\ \midrule\midrule
\multicolumn{1}{c}{\multirow{2}{*}{\textbf{An Type (안 계열)}}} & \textbf{Short-Form} & \begin{tabular}[c]{@{}c@{}}나는 저녁으로 빵을 먹었다.\\  (I ate bread for dinner.)\end{tabular} & \begin{tabular}[c]{@{}c@{}}나는 저녁으로 빵을 {\color{red}안} 먹었다.\\  (I {\color{red}didn’t} eat bread for dinner.)\end{tabular}     \\ \cmidrule(l){2-4} 
\multicolumn{1}{c}{}                                 & \textbf{Long-Form}  & \begin{tabular}[c]{@{}c@{}}나는 저녁으로 빵을 먹었다.\\  (I ate bread for dinner.)\end{tabular} & \begin{tabular}[c]{@{}c@{}}나는 저녁으로 빵을 먹{\color{red}지 않}았다.\\  (I {\color{red}didn’t} eat bread for dinner.)\end{tabular}    \\ \midrule
\multicolumn{1}{c}{\multirow{2}{*}{\textbf{Mot Type (못 계열})}} & \textbf{Short-Form} & \begin{tabular}[c]{@{}c@{}}나는 저녁으로 빵을 먹었다.\\  (I ate bread for dinner.)\end{tabular} & \begin{tabular}[c]{@{}c@{}}나는 저녁으로 빵을 {\color{red}못} 먹었다.\\  (I {\color{red}could not} eat bread for dinner.)\end{tabular}  \\ \cmidrule(l){2-4} 
\multicolumn{1}{c}{}                                 & \textbf{Long-Form}  & \begin{tabular}[c]{@{}c@{}}나는 저녁으로 빵을 먹었다.\\  (I ate bread for dinner.)\end{tabular} & \begin{tabular}[c]{@{}c@{}}나는 저녁으로 빵을 먹{\color{red}지 못}했다.\\  (I {\color{red}could not} eat bread for dinner.)\end{tabular} \\ \midrule
\multicolumn{2}{c}{\textbf{Malda(말다)}}                               & \begin{tabular}[c]{@{}c@{}}저녁을 먹어라.\\  (Eat dinner.)\end{tabular}                     & \begin{tabular}[c]{@{}c@{}}저녁을 먹{\color{red}지 말}아라.\\  ({\color{red}Do not} eat dinner.)\end{tabular}                       \\ \bottomrule
\end{tabular}%
}
\caption{Syntactic negation type and example in Korean}
\label{tab:neg_type}
\end{table*}

%% file: tables/app_sentence_korean.tex
\begin{table*}[htbp]
\centering
\footnotesize
\setlength{\tabcolsep}{3pt}
\renewcommand{\arraystretch}{1.15}

\begin{tabularx}{\textwidth}{@{}
  >{\centering\arraybackslash}p{0.1\textwidth}
  >{\centering\arraybackslash}p{0.15\textwidth}
  >{\raggedright\arraybackslash}X
  >{\raggedright\arraybackslash}p{0.28\textwidth}
@{}}
\toprule
\multicolumn{2}{c}{\textbf{Structure}} & \textbf{Description} & \textbf{Example} \\
\midrule\midrule

\multicolumn{2}{c}{\textbf{Simple Sentence}} &
A sentence that contains a single main(i.e., independent) clause. &
\makecell[l]{나는 집에 갔다.\\(I went home.)} \\
\midrule

\multirow[c]{2}{*}{\textbf{\makecell{Connected\\Sentences}}}
& \textbf{\makecell{Coordinated\\sentences}} &
Two or more main clauses are linked in an equal relationship. &
\makecell[l]{예술은 길고 인생은 짧다.\\(Art is long, but life is short.)} \\
\cmidrule(l){2-4}

& \textbf{With subordinate clause} &
A subordinate clause is combined with a main clause in a dependent relationship. &
\makecell[l]{비가 와서 땅이 젖었다.\\(Because it rained, the ground \\ became wet.)} \\
\midrule

\multirow[c]{5}{*}{\textbf{\makecell{Sentences\\with\\Embedded\\clause}}}
& \textbf{\makecell{Noun\\clause}} &
An embedded clause that functions as a noun (e.g., formed with endings such as -(으)ㅁ or -기). &
\makecell[l]{비가 오기를 기다린다.\\(I am waiting for it to rain.)} \\
\cmidrule(l){2-4}

& \textbf{\makecell{Adnominal\\clause}} &
An embedded clause that modifies a nominal argument (e.g., formed with -(으)ㄴ, -는, -(으)ㄹ, -던). &
\makecell[l]{내가 읽은 책은 재미있다.\\(The book that I read is interesting.)} \\
\cmidrule(l){2-4}

& \textbf{\makecell{Quotation\\clause}} &
An embedded clause that reports/quotes utterance or thought (e.g., formed with -라고, -고). &
\makecell[l]{그가 그녀가 온다고 말했다.\\(He said that she would come.)} \\
\cmidrule(l){2-4}

& \textbf{\makecell{Adverbial\\clause}} &
An embedded clause that functions as an adverbial modifier (e.g., formed with -게, -도록), providing additional information about the main clause. &
\makecell[l]{그는 그 날 날씨가 나빠서 집에\\ 있었다. (He stayed home \\ because the weather is bad.)} \\
\cmidrule(l){2-4}

& \textbf{\makecell{Predicative\\clause}} &
An embedded clause that serves as the predicate of the sentence. &
\makecell[l]{코끼리는 코가 길다.\\(An elephant has a long trunk.)} \\
\bottomrule
\end{tabularx}

\caption{Korean sentence structure.}
\label{tab:sentence_structure}
\end{table*}

%% file: tables/app_corpus.tex
\begin{table}[!t]
\centering
\begin{minipage}{\linewidth}
\centering
\resizebox{\columnwidth}{!}{%
\begin{tabular}{@{}ccc@{}}
\toprule
\textbf{Category} 
& \textbf{Number of Sentences} 
& \textbf{\makecell{Number of \\whitespace-delimited \\Korean word units \\(어절; Eojeol)}} 
\\ \midrule

Spoken-style    
& 683,277 
& 1,025,519,624
\\ 

Written-style
& 2,678,129
& 1,260,946,221
\\ \midrule

\textbf{Total}
& 3,361,406
& 2,286,465,845
\\
\bottomrule
\end{tabular}%
}
\vspace{-0.5\baselineskip}
\caption{Distribution of sentence types in the corpus.}
\label{tab:corpus_sentence_type}
\end{minipage}\\

\vspace{1.1\baselineskip}

\begin{minipage}{\linewidth}
\centering
\resizebox{\columnwidth}{!}{%
\begin{tabular}{@{}ccc@{}}
\toprule
\textbf{Category} 
& \textbf{Number of Sentences} 
& \textbf{Proportion (\%)} 
\\ \midrule

Engineering 
& 61,963 
& 1.84 
\\

Miscellaneous 
& 374,513 
& 11.14 
\\

Daily-life Expressions 
& 3,217 
& 0.10 
\\

Healthcare 
& 186,247 
& 5.54 
\\

Society 
& 1,218,049 
& 36.24 
\\

Industry 
& 600,154 
& 17.85 
\\

Arts \& Entertainment 
& 178,491 
& 5.31 
\\

Humanities 
& 489,046 
& 14.55 
\\

Natural Science 
& 149,095 
& 4.44 
\\

Religion 
& 100,629 
& 2.99 
\\ \midrule

\textbf{Total} 
& 3,361,404 
& 100.00 
\\ \bottomrule
\end{tabular}%
}
\vspace{-0.5\baselineskip}
\caption{Distribution of sentences across topic categories in the corpus.}
\label{tab:corpus_category}
\end{minipage}
\vspace{-1\baselineskip}
\end{table}

%% file: tables/app_inter_annotation.tex
\begin{table}[htbp]
\centering
\resizebox{0.7\columnwidth}{!}{
\begin{tabular}{@{}ccc@{}}

\toprule

& \textbf{\makecell{1st regular\\meeting}} 
& \textbf{\makecell{2nd regular\\meeting}}  
\\ \midrule
\textbf{\makecell{disagreement\\rate}}      
& 12.8\%
& 3.2\%
\\ \bottomrule
\end{tabular}%
}
\caption{Inter-annotation disagreement rates across review stages.}
\label{tab:inter_annotation}
\end{table}

%% file: tables/app_kmmlu_boolq.tex
\begin{table*}[htbp]
\centering
\resizebox{0.93\textwidth}{!}{%
\begin{tabular}{@{}lcccc@{}}
\toprule
& \multicolumn{2}{c}{\textbf{KMMLU}}                                        
& \multicolumn{2}{c}{\textbf{BoolQ}}                               
\\ \cmidrule(l){2-3} \cmidrule(l){4-5}                                     
\textbf{Model}& \multicolumn{1}{c}{\textbf{
\begin{tabular}[c]{@{}c@{}}Negative\end{tabular}}} 
& \multicolumn{1}{c}{\textbf{\begin{tabular}[c]{@{}c@{}}Affirmative \end{tabular}}} 
& \multicolumn{1}{c}{\textbf{Original}}          
& \multicolumn{1}{c}{\textbf{\begin{tabular}[c]{@{}c@{}}Negated \end{tabular}}} 
\\ \midrule\midrule

\textbf{Midm-2.0-Mini-Instruct (2.3B)} & 63.1 & \textbf{63.5} & \textbf{87.3} & 68.2 \\
\textbf{Midm-2.0-Base-Instruct (11.5B)} & 61.7 & \textbf{73.8} & \textbf{92.1} & 77.3 \\
\midrule
\textbf{EXAONE-4.0-1.2B} & \textbf{51.0} & 50.1 & \textbf{51.4} & 48.4 \\
\textbf{EXAONE-4.0-32B} & \textbf{75.4} & 74.2 & \textbf{81.1} & 52.8 \\
\textbf{EXAONE-Deep-2.4B} & 47.8 & \textbf{53.6} & \textbf{51.0} & 48.2 \\
\textbf{EXAONE-Deep-7.8B} & 48.6 & \textbf{56.3} & \textbf{69.6} & 48.9 \\
\textbf{EXAONE-Deep-32B} & 56.6 & \textbf{65.9} & \textbf{85.4} & 68.2 \\
\midrule
\textbf{kanana-1.5-2.1b-base} & 57.4 & \textbf{64.0} & \textbf{54.8} & 48.5 \\
\textbf{kanana-1.5-2.1b-instruct-2505} & 64.5 & \textbf{66.8} & \textbf{66.4} & 51.5 \\
\textbf{kanana-1.5-8b-base} & 66.3 & \textbf{66.5} & \textbf{63.7} & 51.2 \\
\textbf{kanana-1.5-8b-instruct-2505} & 66.5 & \textbf{69.9} & \textbf{80.3} & 56.9 \\
\midrule
\textbf{HyperCLOVAX-SEED-Text-Instruct-0.5B} & 59.9 & \textbf{60.8} & \textbf{53.5} & 48.1 \\
\textbf{HyperCLOVAX-SEED-Text-Instruct-1.5B} & 61.9 & \textbf{63.5} & \textbf{69.3} & 51.0 \\
\textbf{HyperCLOVAX-SEED-Think-14B} & \textbf{75.7} & 74.8 & \textbf{87.0} & 66.6 \\
\midrule
\textbf{A.X-4.0-Light (7.2B)} & 76.2 & \textbf{76.8} & \textbf{93.0} & 74.9 \\
\textbf{A.X-4.0 (72B)} & \textbf{87.7} & 86.5 & \textbf{94.7} & 77.6 \\
\midrule
\textbf{SOLAR-10.7B-v1.0} & 54.0 & \textbf{58.2} & \textbf{50.6} & 48.1 \\
\textbf{SOLAR-10.7B-Instruct-v1.0} & 60.4 & \textbf{62.7} & \textbf{86.0} & 67.7 \\
\midrule
\textbf{Qwen3-0.6B-Base} & \textbf{56.9} & 54.2 & \textbf{52.1} & 49.2 \\
\textbf{Qwen3-0.6B} & 50.0 & \textbf{52.2} & \textbf{50.4} & 48.1 \\
\textbf{Qwen3-1.7B-Base} & 54.9 & \textbf{62.0} & \textbf{61.7} & 48.7 \\
\textbf{Qwen3-1.7B} & 57.2 & \textbf{60.8} & \textbf{51.9} & 48.6 \\
\textbf{Qwen3-4B-Base} & \textbf{70.2} & 69.1 & \textbf{63.7} & 48.9 \\
\textbf{Qwen3-4B} & 59.5 & \textbf{65.9} & \textbf{53.6} & 48.1 \\
\textbf{Qwen3-8B-Base} & \textbf{73.6} & 72.7 & \textbf{57.2} & 48.0 \\
\textbf{Qwen3-8B} & 68.8 & \textbf{70.8} & \textbf{61.2} & 48.9 \\
\textbf{Qwen3-14B-Base} & \textbf{76.3} & 69.0 & \textbf{67.7} & 50.8 \\
\textbf{Qwen3-14B} & \textbf{75.5} & 70.2 & \textbf{64.0} & 54.1 \\
\textbf{Qwen3-32B} & \textbf{79.5} & 77.7 & \textbf{58.8} & 47.6 \\
\midrule
\textbf{Llama-3.2-1B} & \textbf{51.2} & 49.7 & 49.4 & \textbf{49.8} \\
\textbf{Llama-3.2-1B-Instruct} & 50.6 & \textbf{52.4} & \textbf{50.4} & 48.1 \\
\textbf{Llama-3.2-3B} & \textbf{56.7} & 54.2 & \textbf{54.2} & 47.4 \\
\textbf{Llama-3.2-3B-Instruct} & \textbf{59.5} & 57.7 & \textbf{51.5} & 47.9 \\
\textbf{Llama-3.1-8B} & 60.3 & \textbf{60.9} & \textbf{53.0} & 48.2 \\
\textbf{Llama-3.1-8B-Instruct} & 56.3 & \textbf{57.7} & \textbf{54.1} & 48.1 \\
\textbf{Llama-3.1-70B} & \textbf{70.8} & 70.4 & \textbf{67.7} & 49.6 \\
\textbf{Llama-3.1-70B-Instruct} & \textbf{75.0} & 74.7 & \textbf{89.7} & 63.5 \\
\midrule
\textbf{Mistral-7B-v0.3} & 52.4 & \textbf{56.3} & \textbf{58.0} & 48.1 \\
\textbf{Mistral-7B-Instruct-v0.3} & 58.3 & \textbf{59.1} & \textbf{77.1} & 56.0 \\
\textbf{Mistral-Nemo-Base-2407 (12B)} & 54.6 & \textbf{65.3} & \textbf{66.5} & 46.9 \\
\textbf{Mistral-Nemo-Instruct-2407 (12B)} & 62.2 & \textbf{63.1} & \textbf{76.1} & 47.6 \\
\textbf{Mistral-Small-24B-Base-2501} & 66.8 & \textbf{71.9} & \textbf{87.7} & 53.8 \\
\textbf{Mistral-Small-24B-Instruct-2501} & 69.9 & \textbf{72.3} & \textbf{92.7} & 58.1 \\
\bottomrule

\end{tabular}%
}
\caption{Model Performance on KMMLU and BoolQ with and without negation. Bold values indicate the higher performance between the negated and affirmative (or original) versions for each model.}
\label{tab:kmmlu_boolq}
\end{table*}

%% file: tables/app_baseline_result.tex
\begin{table*}[htbp]
\centering
\footnotesize
\resizebox{\textwidth}{!}{%
\begin{tabular}{@{}lcccccccccc@{}}
\toprule
\textbf{} &
  \multicolumn{2}{c}{\textbf{0shot}} &
  \multicolumn{2}{c}{\textbf{1shot}} &
  \multicolumn{2}{c}{\textbf{2shot}} &
  \multicolumn{2}{c}{\textbf{5shot}} &
  \multicolumn{2}{c}{\textbf{10shot}} \\
\textbf{Model} &
  \textbf{cloze} &
  \textbf{symbol} &
  \textbf{cloze} &
  \textbf{symbol} &
  \textbf{cloze} &
  \textbf{symbol} &
  \textbf{cloze} &
  \textbf{symbol} &
  \textbf{cloze} &
  \textbf{symbol} \\ \midrule\midrule

\makecell[l]{\textbf{Midm-2.0-Mini-Instruct}\\\textbf{(2.3B)}} & 60.4 & 55.4 & 70.0 & 65.0 & 77.2 & 69.2 & 84.0 & 68.7 & 87.2 & 70.9 \\
\makecell[l]{\textbf{Midm-2.0-Base-Instruct}\\\textbf{(11.5B)}} & 66.1 & 59.6 & 75.9 & 72.2 & 81.6 & 74.8 & 86.7 & 80.7 & 89.6 & 83.0 \\
\midrule

\textbf{EXAONE-4.0-1.2B} & 32.2 & 26.5 & 44.1 & 29.2 & 49.1 & 41.6 & 60.2 & 43.7 & 65.8 & 44.3 \\
\textbf{EXAONE-4.0-32B} & 51.9 & 81.5 & 65.6 & 86.9 & 72.0 & 88.0 & 80.4 & 91.0 & 83.1 & 92.1 \\
\textbf{EXAONE-Deep-2.4B} & 43.2 & 40.3 & 57.6 & 42.6 & 64.4 & 43.4 & 73.4 & 46.8 & 79.8 & 46.7 \\
\textbf{EXAONE-Deep-7.8B} & 59.3 & 43.7 & 64.9 & 57.3 & 69.6 & 68.2 & 76.4 & 75.4 & 80.1 & 77.3 \\
\textbf{EXAONE-Deep-32B} & 60.1 & 71.7 & 69.5 & 75.5 & 72.7 & 76.6 & 78.7 & 81.1 & 82.4 & 84.2 \\
\midrule

\textbf{kanana-1.5-2.1b-base} & 50.1 & 45.2 & 65.0 & 44.0 & 72.1 & 47.0 & 80.3 & 46.3 & 85.4 & 49.5 \\
\makecell[l]{\textbf{kanana-1.5-2.1b-}\\\textbf{instruct-2505}} & 64.6 & 57.5 & 78.3 & 60.8 & 82.1 & 61.9 & 87.7 & 58.2 & 90.3 & 57.3 \\
\textbf{kanana-1.5-8b-base} & 45.8 & 43.6 & 60.0 & 62.0 & 70.3 & 62.1 & 78.8 & 67.7 & 82.7 & 71.8 \\
\makecell[l]{\textbf{kanana-1.5-8b-}\\\textbf{instruct-2505}} & 55.9 & 56.8 & 70.6 & 67.5 & 77.5 & 74.4 & 87.0 & 81.1 & 89.4 & 82.4 \\
\midrule

\makecell[l]{\textbf{HyperCLOVAX-SEED-}\\\textbf{Text-Instruct-0.5B}} & 38.0 & 52.7 & 46.6 & 37.0 & 52.0 & 41.9 & 61.1 & 25.6 & 54.7 & 45.8 \\
\makecell[l]{\textbf{HyperCLOVAX-SEED-}\\\textbf{Text-Instruct-1.5B}} & 38.5 & 46.7 & 53.8 & 46.4 & 60.2 & 52.6 & 68.7 & 53.9 & 74.1 & 54.0 \\
\makecell[l]{\textbf{HyperCLOVAX-SEED-}\\\textbf{Think-14B}} & 47.5 & 78.1 & 70.1 & 88.1 & 77.1 & 89.9 & 83.3 & 92.0 & 86.4 & 92.1 \\
\midrule
\textbf{A.X-4.0-Light (7.2B)} & 71.3 & 85.0 & 72.1 & 86.1 & 77.5 & 88.3 & 84.9 & 88.4 & 86.8 & 93.3 \\
\textbf{A.X-4.0 (72B)} & {\color{red}76.3} & 95.8 & {\color{red}84.4} & 95.1 & {\color{red}87.9} & 95.1 & {\color{red}90.6} & 91.1 & 90.3 & 95.9 \\
\midrule

\textbf{SOLAR-10.7B-v1.0} & 42.2 & 42.1 & 58.8 & 56.7 & 65.8 & 58.0 & 74.6 & 62.1 & 80.3 & 62.6 \\
\makecell[l]{\textbf{SOLAR-10.7B-}\\\textbf{Instruct-v1.0}} & 64.8 & 54.8 & 77.4 & 61.2 & 81.0 & 63.6 & 85.0 & 67.0 & 86.4 & 68.1 \\
\midrule
\textbf{Qwen3-0.6B-Base} & 41.4 & 51.6 & 60.7 & 49.6 & 65.8 & 53.0 & 73.2 & 53.8 & 77.6 & 58.4 \\
\textbf{Qwen3-0.6B} & 39.1 & 64.6 & 48.8 & 49.7 & 51.1 & 47.6 & 56.4 & 54.1 & 61.7 & 56.2 \\
\textbf{Qwen3-1.7B-Base} & 64.9 & 64.7 & 73.7 & 72.8 & 78.9 & 79.7 & 86.5 & 81.9 & 88.8 & 84.1 \\
\textbf{Qwen3-1.7B} & 54.8 & 76.1 & 61.3 & 78.7 & 67.3 & 82.7 & 77.3 & 86.2 & 82.7 & 84.5 \\
\textbf{Qwen3-4B-Base} & 67.8 & 82.2 & 76.3 & 85.5 & 80.4 & 87.1 & 87.2 & 90.0 & 89.5 & 91.2 \\
\textbf{Qwen3-4B} & 47.7 & 80.5 & 63.6 & 85.7 & 69.7 & 87.4 & 78.5 & 89.3 & 82.7 & 89.8 \\
\textbf{Qwen3-8B-Base} & 60.5 & 85.5 & 72.8 & 88.8 & 79.7 & 90.3 & 87.5 & 92.5 & 90.5 & 92.9 \\
\textbf{Qwen3-8B} & 47.7 & 74.8 & 64.1 & 83.0 & 70.6 & 86.0 & 81.6 & 89.3 & 85.5 & 92.0 \\
\textbf{Qwen3-14B-Base} & 65.8 & 88.6 & 76.9 & 90.9 & 80.3 & 92.2 & 86.3 & 93.4 & 89.5 & 94.1 \\
\textbf{Qwen3-14B} & 47.9 & 90.6 & 66.3 & 92.5 & 73.6 & 92.5 & 83.1 & 93.0 & 86.9 & 94.0 \\
\textbf{Qwen3-32B} & 57.6 & 94.2 & 75.2 & 94.5 & 82.5 & 94.8 & 88.8 & 95.8 & 89.8 & 96.9 \\
\midrule

\textbf{Llama-3.2-1B} & 38.4 & 26.2 & 47.7 & 24.0 & 52.5 & 24.9 & 61.7 & 25.3 & 69.2 & 26.2 \\
\textbf{Llama-3.2-1B-Instruct} & 37.2 & 33.6 & 46.6 & 29.3 & 51.7 & 34.8 & 61.2 & 34.9 & 67.5 & 33.4 \\
\textbf{Llama-3.2-3B} & 45.2 & 52.0 & 58.7 & 50.6 & 64.0 & 58.3 & 74.0 & 62.7 & 78.3 & 64.5 \\
\textbf{Llama-3.2-3B-Instruct} & 41.1 & 61.8 & 46.5 & 60.8 & 51.2 & 64.4 & 64.7 & 69.0 & 73.5 & 69.7 \\
\textbf{Llama-3.1-8B} & 40.4 & 53.6 & 54.7 & 64.3 & 63.2 & 70.7 & 74.9 & 74.7 & 81.3 & 77.6 \\
\textbf{Llama-3.1-8B-Instruct} & 37.4 & 57.2 & 52.8 & 66.8 & 61.9 & 68.0 & 74.1 & 71.3 & 80.5 & 72.4 \\
\textbf{Llama-3.1-70B} & 57.5 & 76.6 & 70.8 & 85.2 & 75.8 & 88.3 & 83.0 & 91.0 & 87.8 & 92.4 \\
\textbf{Llama-3.1-70B-Instruct} & 50.1 & 81.0 & 69.5 & 86.9 & 76.4 & 88.9 & 85.5 & 92.3 & 89.6 & 93.0 \\
\midrule

\textbf{Mistral-7B-v0.3} & 51.5 & 46.9 & 65.2 & 55.6 & 72.1 & 67.3 & 79.3 & 77.6 & 82.9 & 77.5 \\
\textbf{Mistral-7B-Instruct-v0.3} & 65.8 & 73.0 & 76.5 & 71.9 & 80.0 & 69.9 & 84.7 & 67.7 & 86.3 & 65.3 \\
\makecell[l]{\textbf{Mistral-Nemo-}}\\\textbf{Base-2407 (12B)} & 49.7 & 51.2 & 69.9 & 69.9 & 79.5 & 76.2 & 87.0 & 76.6 & 90.7 & 74.8 \\
\makecell[l]{\textbf{Mistral-Nemo-}}\\\textbf{Instruct-2407 (12B)} & 55.6 & 65.1 & 73.2 & 75.1 & 80.0 & 78.3 & 87.1 & 82.4 & 90.7 & 81.6 \\
\makecell[l]{\textbf{Mistral-Small-24B-}}\\\textbf{Base-2501} & 61.3 & 71.6 & 75.1 & 83.9 & 82.3 & 89.3 & 87.7 & 94.8 & 90.3 & 95.2 \\
\makecell[l]{\textbf{Mistral-Small-24B-}}\\\textbf{Instruct-2501} & 60.4 & 81.9 & 77.2 & 87.7 & 84.1 & 91.8 & 90.1 & 95.8 & {\color{red}91.5} & 96.1 \\
\midrule

\textbf{claude-haiku-4-5-20251001} & - & 92.8 & - & 94.0 & - & 95.4 & - & 96.2 & - & 96.6 \\
\textbf{claude-sonnet-4-5-20250929} & - & {\color{red}98.1} & - & {\color{red}98.6} & - & {\color{red}98.4} & - & {\color{red}97.6} & - & {\color{red}97.0} \\
\midrule

\textbf{gpt-4.1-mini} & - & 83.6 & - & 88.1 & - & 88.1 & - & 90.8 & - & 92.4 \\
\textbf{gpt-4.1} & - & 92.2 & - & 92.8 & - & 93.8 & - & 94.5 & - & 95.0 \\
\bottomrule

\end{tabular}%
}
\caption{Model Performance on zero-shot, few-shot setting on \bench. Few-shot results are averaged over 3 random seeds (1234, 308, 1028). Red text indicates the model with highest performance in each setting.}
\label{tab:baseline_result}
\end{table*}

%% file: tables/app_sft_cloze_result.tex
\begin{table*}[htbp]
\centering
\footnotesize
\renewcommand{\arraystretch}{1.3}
\resizebox{\textwidth}{!}{%
\begin{tabular}{lcccccccccc}
\toprule

& \multicolumn{2}{c}{\textbf{\bench}} &
\multicolumn{2}{c}{\textbf{KMMLU}} &
\multicolumn{2}{c}{\textbf{BoolQ}} \\
\cmidrule(l){2-3} \cmidrule(l){4-5} \cmidrule(l){6-7}
\textbf{Model} & \textbf{cloze} & \textbf{symbol} & \textbf{negative} & \textbf{affirmative} & \textbf{original} & \textbf{negated} \\
\midrule\midrule

\textbf{Midm-2.0-Base-Instruct}  & \makecell{84.7 (+18.6)} & \makecell{66.7 (+7.1)} & \makecell{61.7 (+0.0)} & \makecell{73.5 (-0.3)} & \makecell{92.0 (-0.1)} & \makecell{77.6 (+0.3)} \\
\textbf{Midm-2.0-Mini-Instruct}  & \makecell{91.0 (+30.6)} & \makecell{66.9 (+11.5)} & \makecell{64.3 (+1.2)} & \makecell{64.5 (+1.0)} & \makecell{85.3 (-2.0)} & \makecell{59.0 (-9.2)} \\
\midrule

\textbf{EXAONE-4.0-1.2B}  & \makecell{67.4 (+35.2)} & \makecell{31.7 (+5.2)} & \makecell{50.9 (-0.1)} & \makecell{49.9 (-0.2)} & \makecell{56.8 (+5.4)} & \makecell{49.2 (+0.8)} \\
\textbf{EXAONE-Deep-2.4B}  & \makecell{67.8 (+24.6)} & \makecell{49.8 (+9.5)} & \makecell{48.2 (+0.4)} & \makecell{53.4 (-0.2)} & \makecell{50.9 (-0.1)} & \makecell{48.1 (-0.1)} \\
\textbf{EXAONE-Deep-7.8B}  & \makecell{81.7 (+22.4)} & \makecell{60.2 (+16.5)} & \makecell{49.3 (+0.7)} & \makecell{56.2 (-0.1)} & \makecell{71.7 (+2.1)} & \makecell{50.9 (+2.0)} \\
\midrule

\textbf{kanana-1.5-2.1b-base}  & \makecell{92.7 (+42.6)} & \makecell{62.2 (+17.0)} & \makecell{56.8 (-0.6)} & \makecell{62.7 (-1.3)} & \makecell{63.0 (+8.2)} & \makecell{49.8 (+1.3)} \\
\textbf{kanana-1.5-2.1b-instruct-2505}  & \makecell{92.1 (+27.5)} & \makecell{67.5 (+10.0)} & \makecell{65.0 (+0.5)} & \makecell{66.9 (+0.1)} & \makecell{71.2 (+4.8)} & \makecell{54.6 (+3.1)} \\
\textbf{kanana-1.5-8b-base}  & \makecell{94.1 (+48.3)} & \makecell{60.3 (+16.7)} & \makecell{62.0 (-4.3)} & \makecell{62.0 (-4.5)} & \makecell{69.2 (+5.5)} & \makecell{51.7 (+0.5)} \\
\textbf{kanana-1.5-8b-instruct-2505}  & \makecell{91.4 (+35.5)} & \makecell{68.5 (+11.7)} & \makecell{67.2 (+0.7)} & \makecell{70.2 (+0.3)} & \makecell{83.6 (+3.3)} & \makecell{61.2 (+4.3)} \\
\midrule

\textbf{HyperCLOVAX-SEED-Text-Instruct-0.5B}  & \makecell{69.2 (+31.2)} & \makecell{60.3 (+7.6)} & \makecell{61.0 (+1.1)} & \makecell{61.2 (+0.4)} & \makecell{56.3 (+2.8)} & \makecell{48.3 (+0.2)} \\
\textbf{HyperCLOVAX-SEED-Text-Instruct-1.5B}  & \makecell{70.5 (+32.0)} & \makecell{49.6 (+2.9)} & \makecell{62.0 (+0.1)} & \makecell{63.4 (-0.1)} & \makecell{70.6 (+1.3)} & \makecell{50.9 (-0.1)} \\
\textbf{HyperCLOVAX-SEED-Think-14B}  & \makecell{86.4 (+38.9)} & \makecell{86.4 (+8.3)} & \makecell{75.9 (+0.2)} & \makecell{75.0 (+0.2)} & \makecell{87.6 (+0.6)} & \makecell{67.4 (+0.8)} \\
\midrule

\textbf{A.X-4.0-Light (7.2B)}  & \makecell{85.1 (+13.8)} & \makecell{88.9 (+3.9)} & \makecell{76.3 (+0.1)} & \makecell{76.6 (-0.2)} & \makecell{92.9 (-0.1)} & \makecell{74.9 (+0.0)} \\
\midrule

\textbf{SOLAR-10.7B-v1.0}  & \makecell{97.6 (+55.4)} & \makecell{79.3 (+37.2)} & \makecell{54.9 (+0.9)} & \makecell{58.4 (+0.2)} & \makecell{58.3 (+7.7)} & \makecell{50.6 (+2.5)} \\
\textbf{SOLAR-10.7B-Instruct-v1.0}  & \makecell{97.2 (+32.4)} & \makecell{74.5 (+19.7)} & \makecell{62.0 (+1.6)} & \makecell{62.5 (-0.2)} & \makecell{88.1 (+2.1)} & \makecell{75.0 (+7.3)} \\
\midrule

\textbf{Qwen3-0.6B-Base}  & \makecell{78.0 (+36.6)} & \makecell{59.0 (+7.4)} & \makecell{57.5 (+0.6)} & \makecell{54.0 (-0.2)} & \makecell{52.6 (+0.5)} & \makecell{51.3 (+2.1)} \\
\textbf{Qwen3-0.6B}  & \makecell{73.7 (+34.6)} & \makecell{65.5 (+0.9)} & \makecell{50.1 (+0.1)} & \makecell{52.0 (-0.2)} & \makecell{50.4 (+0.0)} & \makecell{48.0 (-0.1)} \\
\textbf{Qwen3-1.7B-Base}  & \makecell{86.0 (+21.1)} & \makecell{66.0 (+1.3)} & \makecell{55.8 (+0.9)} & \makecell{62.2 (+0.2)} & \makecell{63.9 (+2.2)} & \makecell{48.8 (+0.1)} \\
\textbf{Qwen3-1.7B}  & \makecell{74.8 (+20.0)} & \makecell{78.2 (+2.1)} & \makecell{57.8 (+0.6)} & \makecell{60.7 (-0.1)} & \makecell{51.9 (+0.0)} & \makecell{48.4 (-0.2)} \\
\textbf{Qwen3-4B-Base}  & \makecell{89.3 (+21.5)} & \makecell{87.2 (+5.0)} & \makecell{70.5 (+0.3)} & \makecell{68.9 (-0.2)} & \makecell{68.8 (+5.1)} & \makecell{50.8 (+1.9)} \\
\textbf{Qwen3-4B}  & \makecell{79.4 (+31.7)} & \makecell{86.9 (+6.4)} & \makecell{61.2 (+1.7)} & \makecell{66.1 (+0.2)} & \makecell{52.8 (-0.8)} & \makecell{47.9 (-0.2)} \\
\textbf{Qwen3-8B-Base}  & \makecell{89.9 (+29.4)} & \makecell{91.2 (+5.7)} & \makecell{74.0 (+0.4)} & \makecell{72.6 (-0.1)} & \makecell{58.5 (+1.3)} & \makecell{47.8 (-0.2)} \\
\textbf{Qwen3-8B}  & \makecell{87.3 (+39.6)} & \makecell{87.1 (+12.3)} & \makecell{69.5 (+0.7)} & \makecell{70.8 (+0.0)} & \makecell{62.7 (+1.5)} & \makecell{48.9 (+0.0)} \\
\textbf{Qwen3-14B-Base}  & \makecell{89.9 (+24.1)} & \makecell{93.1 (+4.5)} & \makecell{76.2 (-0.1)} & \makecell{69.4 (+0.4)} & \makecell{70.9 (+3.2)} & \makecell{51.4 (+0.6)} \\
\textbf{Qwen3-14B}  & \makecell{86.2 (+38.3)} & \makecell{92.9 (+2.3)} & \makecell{75.5 (+0.0)} & \makecell{70.1 (-0.1)} & \makecell{72.0 (+8.0)} & \makecell{56.6 (+2.5)} \\
\midrule

\textbf{Llama-3.2-1B}  & \makecell{71.7 (+33.3)} & \makecell{26.1 (-0.1)} & \makecell{51.0 (-0.2)} & \makecell{50.0 (+0.3)} & \makecell{51.1 (+1.7)} & \makecell{48.8 (-1.0)} \\
\textbf{Llama-3.2-1B-Instruct}  & \makecell{68.7 (+31.5)} & \makecell{36.8 (+3.2)} & \makecell{51.0 (+0.4)} & \makecell{52.2 (-0.2)} & \makecell{50.4 (+0.0)} & \makecell{48.0 (-0.1)} \\
\textbf{Llama-3.2-3B}  & \makecell{85.7 (+40.5)} & \makecell{58.5 (+6.5)} & \makecell{57.0 (+0.3)} & \makecell{54.1 (-0.1)} & \makecell{58.6 (+4.4)} & \makecell{48.0 (+0.6)} \\
\textbf{Llama-3.2-3B-Instruct}  & \makecell{81.9 (+40.8)} & \makecell{67.0 (+5.2)} & \makecell{59.4 (-0.1)} & \makecell{58.1 (+0.4)} & \makecell{54.0 (+2.5)} & \makecell{48.3 (+0.4)} \\
\textbf{Llama-3.1-8B}  & \makecell{91.8 (+51.4)} & \makecell{70.4 (+16.8)} & \makecell{60.8 (+0.5)} & \makecell{61.1 (+0.2)} & \makecell{59.3 (+6.3)} & \makecell{48.8 (+0.6)} \\
\textbf{Llama-3.1-8B-Instruct}  & \makecell{90.1 (+52.7)} & \makecell{73.7 (+16.5)} & \makecell{58.4 (+2.1)} & \makecell{59.4 (+1.7)} & \makecell{65.3 (+11.2)} & \makecell{47.9 (-0.2)} \\
\midrule

\textbf{Mistral-7B-v0.3}  & \makecell{97.4 (+45.9)} & \makecell{80.1 (+33.2)} & \makecell{52.5 (+0.1)} & \makecell{55.5 (-0.8)} & \makecell{65.5 (+7.5)} & \makecell{52.3 (+4.2)} \\
\textbf{Mistral-7B-Instruct-v0.3}  & \makecell{97.0 (+31.2)} & \makecell{91.4 (+18.4)} & \makecell{58.2 (-0.1)} & \makecell{57.9 (-1.2)} & \makecell{79.5 (+2.4)} & \makecell{65.5 (+9.5)} \\
\textbf{Mistral-Nemo-Base-2407 (12B)}  & \makecell{95.4 (+45.7)} & \makecell{69.7 (+18.5)} & \makecell{56.7 (+2.1)} & \makecell{65.9 (+0.6)} & \makecell{71.3 (+4.8)} & \makecell{46.4 (-0.5)} \\
\textbf{Mistral-Nemo-Instruct-2407 (12B)}  & \makecell{93.8 (+38.2)} & \makecell{81.2 (+16.1)} & \makecell{63.0 (+0.8)} & \makecell{63.7 (+0.6)} & \makecell{79.3 (+3.2)} & \makecell{46.7 (-0.9)} \\
\midrule

\textbf{Mean $\Delta$ over Baseline} & 34.2 & 10.5 & 0.4 & -0.1 & 3.0 & 0.9 \\

\bottomrule
\end{tabular}%
}
\caption{Model performance on \bench, KMMLU, and KoBest BoolQ after fine-tuning targeting the cloze format. Values in parentheses indicate the performance gain relative to the baseline results.}
\label{tab:sft_cloze1}
\end{table*}

\begin{table*}[htbp]
\centering
\footnotesize
\renewcommand{\arraystretch}{1.3}
\resizebox{\textwidth}{!}{%
\begin{tabular}{lcccc}
\toprule

 & \multicolumn{2}{c}{\textbf{ARC}} & \textbf{Hellaswag} & \textbf{Winogrande} \\
\cmidrule(l){2-3}
\textbf{Model} & \textbf{easy} & \textbf{challenge} & & \\
 \midrule \midrule

\textbf{Midm-2.0-Mini-Instruct (2.3B)}  & \makecell{79.3 (+0.8)} & \makecell{50.2 (-0.2)} & \makecell{70.3 (+0.1)} & \makecell{66.7 (-0.1)} \\
\textbf{Midm-2.0-Base-Instruct (11.5B)}  & \makecell{84.6 (+0.1)} & \makecell{62.6 (+1.1)} & \makecell{80.5 (+0.1)} & \makecell{73.6 (+0.2)} \\
\midrule

\textbf{EXAONE-4.0-1.2B}  & \makecell{33.2 (+0.2)} & \makecell{23.7 (-1.0)} & \makecell{32.2 (+1.1)} & \makecell{49.3 (-0.8)} \\
\textbf{EXAONE-Deep-2.4B}  & \makecell{56.9 (+2.6)} & \makecell{38.2 (-0.2)} & \makecell{55.8 (+0.6)} & \makecell{54.0 (-0.7)} \\
\textbf{EXAONE-Deep-7.8B}  & \makecell{66.9 (+1.9)} & \makecell{45.2 (+1.4)} & \makecell{64.3 (+0.7)} & \makecell{57.5 (-0.3)} \\
\midrule

\textbf{kanana-1.5-2.1b-base}  & \makecell{76.9 (-1.4)} & \makecell{50.0 (-0.6)} & \makecell{66.1 (+0.1)} & \makecell{65.6 (+1.7)} \\
\textbf{kanana-1.5-2.1b-instruct-2505}  & \makecell{76.2 (+2.3)} & \makecell{51.0 (+0.9)} & \makecell{67.1 (+0.2)} & \makecell{62.7 (-0.3)} \\
\textbf{kanana-1.5-8b-base}  & \makecell{82.3 (+1.0)} & \makecell{55.1 (+1.0)} & \makecell{77.5 (+0.1)} & \makecell{72.6 (-0.4)} \\
\textbf{kanana-1.5-8b-instruct-2505}  & \makecell{34.7 (+0.6)} & \makecell{29.9 (+0.0)} & \makecell{78.8 (+0.1)} & \makecell{72.5 (+0.7)} \\
\midrule

\textbf{HyperCLOVAX-SEED-Text-Instruct-0.5B}  & \makecell{65.2 (-0.2)} & \makecell{37.5 (+0.2)} & \makecell{52.1 (-0.1)} & \makecell{54.9 (+0.3)} \\
\textbf{HyperCLOVAX-SEED-Text-Instruct-1.5B}  & \makecell{66.2 (+0.1)} & \makecell{44.3 (+0.4)} & \makecell{60.9 (+0.0)} & \makecell{57.1 (-0.3)} \\
\textbf{HyperCLOVAX-SEED-Think-14B}  & \makecell{75.4 (+0.5)} & \makecell{54.7 (+0.0)} & \makecell{79.4 (+0.0)} & \makecell{71.2 (+0.0)} \\
\midrule

\textbf{A.X-4.0-Light (7.2B)}  & \makecell{75.3 (+0.8)} & \makecell{56.2 (+0.3)} & \makecell{77.2 (+0.1)} & \makecell{71.3 (+0.2)} \\
\midrule

\textbf{SOLAR-10.7B-v1.0}  & \makecell{77.6 (-0.4)} & \makecell{54.2 (-1.3)} & \makecell{83.3 (+0.2)} & \makecell{74.8 (+0.1)} \\
\textbf{SOLAR-10.7B-Instruct-v1.0}  & \makecell{81.4 (+0.3)} & \makecell{61.7 (-0.5)} & \makecell{86.5 (-0.1)} & \makecell{77.0 (+1.9)} \\
\midrule

\textbf{Qwen3-0.6B-Base}  & \makecell{59.9 (+2.0)} & \makecell{37.9 (-0.1)} & \makecell{53.6 (+0.0)} & \makecell{58.8 (-0.3)} \\
\textbf{Qwen3-0.6B}  & \makecell{31.9 (+0.5)} & \makecell{27.8 (-0.4)} & \makecell{47.1 (-0.1)} & \makecell{55.6 (-0.1)} \\
\textbf{Qwen3-1.7B-Base}  & \makecell{70.2 (+1.9)} & \makecell{45.6 (+0.7)} & \makecell{66.5 (+0.0)} & \makecell{64.8 (+0.4)} \\
\textbf{Qwen3-1.7B}  & \makecell{69.4 (-0.3)} & \makecell{41.6 (-1.2)} & \makecell{60.6 (+0.2)} & \makecell{61.2 (-0.5)} \\
\textbf{Qwen3-4B-Base}  & \makecell{77.2 (+1.4)} & \makecell{52.4 (+0.9)} & \makecell{73.6 (-0.1)} & \makecell{70.9 (+0.4)} \\
\textbf{Qwen3-4B}  & \makecell{78.9 (+0.4)} & \makecell{54.0 (+0.3)} & \makecell{68.6 (+0.2)} & \makecell{66.5 (+0.6)} \\
\textbf{Qwen3-8B-Base}  & \makecell{80.4 (+0.1)} & \makecell{57.3 (+0.6)} & \makecell{78.7 (+0.1)} & \makecell{72.4 (-0.3)} \\
\textbf{Qwen3-8B}  & \makecell{80.9 (+0.0)} & \makecell{56.0 (-0.3)} & \makecell{75.1 (+0.2)} & \makecell{68.0 (+0.4)} \\
\textbf{Qwen3-14B-Base}  & \makecell{82.0 (+0.0)} & \makecell{59.1 (+0.1)} & \makecell{81.4 (+0.0)} & \makecell{73.8 (-0.2)} \\
\textbf{Qwen3-14B}  & \makecell{83.2 (+0.4)} & \makecell{60.9 (+0.7)} & \makecell{78.9 (+0.0)} & \makecell{73.2 (+0.2)} \\
\midrule

\textbf{Llama-3.2-1B}  & \makecell{61.5 (-0.3)} & \makecell{37.4 (+0.3)} & \makecell{64.3 (+0.2)} & \makecell{60.7 (+0.0)} \\
\textbf{Llama-3.2-1B-Instruct}  & \makecell{63.7 (-0.1)} & \makecell{38.1 (+0.4)} & \makecell{61.6 (+0.0)} & \makecell{62.0 (+0.7)} \\
\textbf{Llama-3.2-3B}  & \makecell{72.2 (+0.4)} & \makecell{46.5 (+0.3)} & \makecell{74.1 (+0.0)} & \makecell{69.6 (-0.1)} \\
\textbf{Llama-3.2-3B-Instruct}  & \makecell{71.5 (+0.4)} & \makecell{46.4 (+0.2)} & \makecell{71.8 (+0.2)} & \makecell{69.2 (+0.4)} \\
\textbf{Llama-3.1-8B}  & \makecell{82.4 (-0.2)} & \makecell{55.5 (+0.7)} & \makecell{79.3 (+0.0)} & \makecell{73.3 (-1.1)} \\
\textbf{Llama-3.1-8B-Instruct}  & \makecell{78.0 (+0.8)} & \makecell{54.8 (+0.6)} & \makecell{79.6 (-0.1)} & \makecell{74.2 (+0.1)} \\
\midrule

\textbf{Mistral-7B-v0.3}  & \makecell{80.0 (-0.1)} & \makecell{54.3 (-0.1)} & \makecell{80.7 (+0.1)} & \makecell{73.8 (-0.2)} \\
\textbf{Mistral-7B-Instruct-v0.3}  & \makecell{81.2 (-0.6)} & \makecell{60.1 (-0.1)} & \makecell{83.2 (-0.1)} & \makecell{76.1 (+1.3)} \\
\textbf{Mistral-Nemo-Base-2407 (12B)}  & \makecell{83.8 (+0.3)} & \makecell{60.2 (+0.6)} & \makecell{82.9 (-0.1)} & \makecell{76.8 (-0.1)} \\
\textbf{Mistral-Nemo-Instruct-2407 (12B)}  & \makecell{80.9 (+0.3)} & \makecell{59.4 (-0.2)} & \makecell{82.4 (+0.0)} & \makecell{76.9 (+0.0)} \\

\midrule

\textbf{Mean $\Delta$ over Baseline} & 0.5 & 0.2 & 0.1 & 0.1 \\

 \bottomrule
\end{tabular}%
}
\caption{Model performance on ARC, Hellaswag, and Winogrande after fine-tuning targeting the cloze format. Values in parentheses indicate the performance gain relative to the baseline results.}
\label{tab:sft_cloze2}
\end{table*}

%% file: tables/app_sft_symbol_result.tex
\begin{table*}[htbp]
\centering
\footnotesize
\renewcommand{\arraystretch}{1.3}
\resizebox{\textwidth}{!}{%
\begin{tabular}{lcccccccccc}
\toprule

& \multicolumn{2}{c}{\textbf{\bench}} &
\multicolumn{2}{c}{\textbf{KMMLU}} &
\multicolumn{2}{c}{\textbf{BoolQ}} \\
\cmidrule(l){2-3} \cmidrule(l){4-5} \cmidrule(l){6-7}
\textbf{Model} & \textbf{cloze} & \textbf{symbol} & \textbf{negative} & \textbf{affirmative} & \textbf{original} & \textbf{negated} \\
\midrule\midrule

\textbf{Midm-2.0-Mini-Instruct (2.3B)}  & \makecell{70.0 (+9.6)} & \makecell{92.6 (+37.2)} & \makecell{66.0 (+2.9)} & \makecell{65.2 (+1.7)} & \makecell{87.7 (+0.4)} & \makecell{67.2 (-1.0)} \\
\textbf{Midm-2.0-Base-Instruct (11.5B)}  & \makecell{67.1 (+1.0)} & \makecell{79.6 (+20.0)} & \makecell{61.7 (+0.0)} & \makecell{73.7 (-0.1)} & \makecell{92.2 (+0.1)} & \makecell{77.3 (+0.0)} \\
\midrule

\textbf{EXAONE-4.0-1.2B}  & \makecell{34.0 (+1.8)} & \makecell{85.0 (+58.5)} & \makecell{51.5 (+0.5)} & \makecell{50.3 (+0.2)} & \makecell{51.4 (+0.0)} & \makecell{48.1 (-0.3)} \\
\textbf{EXAONE-Deep-2.4B}  & \makecell{44.5 (+1.3)} & \makecell{60.2 (+19.9)} & \makecell{48.6 (+0.8)} & \makecell{53.6 (+0.0)} & \makecell{50.6 (-0.4)} & \makecell{48.1 (-0.1)} \\
\textbf{EXAONE-Deep-7.8B}  & \makecell{64.4 (+5.1)} & \makecell{93.2 (+49.5)} & \makecell{48.8 (+0.2)} & \makecell{56.4 (+0.1)} & \makecell{71.9 (+2.3)} & \makecell{49.9 (+1.0)} \\
\midrule

\textbf{kanana-1.5-2.1b-base}  & \makecell{64.3 (+14.2)} & \makecell{92.6 (+47.4)} & \makecell{59.8 (+2.4)} & \makecell{62.9 (-1.1)} & \makecell{59.9 (+5.1)} & \makecell{49.3 (+0.8)} \\
\textbf{kanana-1.5-2.1b-instruct-2505}  & \makecell{77.2 (+12.6)} & \makecell{97.8 (+40.3)} & \makecell{65.5 (+1.0)} & \makecell{66.7 (-0.1)} & \makecell{67.7 (+1.3)} & \makecell{52.7 (+1.2)} \\
\textbf{kanana-1.5-8b-base}  & \makecell{69.0 (+23.2)} & \makecell{97.4 (+53.8)} & \makecell{68.6 (+2.3)} & \makecell{68.5 (+2.0)} & \makecell{66.1 (+2.4)} & \makecell{51.6 (+0.4)} \\
\textbf{kanana-1.5-8b-instruct-2505}  & \makecell{74.1 (+18.2)} & \makecell{97.4 (+40.6)} & \makecell{68.1 (+1.6)} & \makecell{69.3 (-0.6)} & \makecell{79.8 (-0.5)} & \makecell{56.3 (-0.6)} \\
\midrule

\textbf{HyperCLOVAX-SEED-Text-Instruct-0.5B}  & \makecell{38.3 (+0.3)} & \makecell{89.5 (+36.8)} & \makecell{62.6 (+2.7)} & \makecell{61.2 (+0.4)} & \makecell{54.6 (+1.1)} & \makecell{47.8 (-0.3)} \\
\textbf{HyperCLOVAX-SEED-Text-Instruct-1.5B}  & \makecell{40.5 (+2.0)} & \makecell{87.0 (+40.3)} & \makecell{62.8 (+0.9)} & \makecell{63.6 (+0.1)} & \makecell{70.3 (+1.0)} & \makecell{51.0 (+0.0)} \\
\textbf{HyperCLOVAX-SEED-Think-14B}  & \makecell{54.5 (+7.0)} & \makecell{98.0 (+19.9)} & \makecell{75.7 (+0.0)} & \makecell{74.6 (-0.2)} & \makecell{87.0 (+0.0)} & \makecell{66.7 (+0.1)} \\
\midrule

\textbf{A.X-4.0-Light (7.2B)}  & \makecell{73.7 (+2.4)} & \makecell{95.3 (+10.3)} & \makecell{76.6 (+0.4)} & \makecell{76.5 (-0.3)} & \makecell{93.1 (+0.1)} & \makecell{74.4 (-0.5)} \\
\midrule

\textbf{SOLAR-10.7B-v1.0}  & \makecell{52.5 (+10.3)} & \makecell{99.5 (+57.4)} & \makecell{53.2 (-0.8)} & \makecell{55.2 (-3.0)} & \makecell{53.7 (+3.1)} & \makecell{48.6 (+0.5)} \\
\textbf{SOLAR-10.7B-Instruct-v1.0}  & \makecell{77.5 (+12.7)} & \makecell{98.8 (+44.0)} & \makecell{61.1 (+0.7)} & \makecell{61.5 (-1.2)} & \makecell{80.7 (-5.3)} & \makecell{63.1 (-4.6)} \\
\midrule

\textbf{Qwen3-0.6B-Base}  & \makecell{45.8 (+4.4)} & \makecell{85.4 (+33.8)} & \makecell{57.4 (+0.5)} & \makecell{54.4 (+0.2)} & \makecell{52.6 (+0.5)} & \makecell{49.2 (+0.0)} \\
\textbf{Qwen3-0.6B}  & \makecell{39.6 (+0.5)} & \makecell{90.3 (+25.7)} & \makecell{50.1 (+0.1)} & \makecell{52.5 (+0.3)} & \makecell{50.2 (-0.2)} & \makecell{48.1 (+0.0)} \\
\textbf{Qwen3-1.7B-Base}  & \makecell{69.0 (+4.1)} & \makecell{87.5 (+22.8)} & \makecell{54.2 (-0.7)} & \makecell{61.8 (-0.2)} & \makecell{64.7 (+3.0)} & \makecell{48.6 (-0.1)} \\
\textbf{Qwen3-1.7B}  & \makecell{55.7 (+0.9)} & \makecell{89.3 (+13.2)} & \makecell{57.1 (-0.1)} & \makecell{59.9 (-0.9)} & \makecell{51.0 (-0.9)} & \makecell{48.1 (-0.5)} \\
\textbf{Qwen3-4B-Base}  & \makecell{72.9 (+5.1)} & \makecell{96.2 (+14.0)} & \makecell{70.2 (+0.0)} & \makecell{68.9 (-0.2)} & \makecell{70.0 (+6.3)} & \makecell{51.0 (+2.1)} \\
\textbf{Qwen3-4B}  & \makecell{47.7 (+0.0)} & \makecell{93.5 (+13.0)} & \makecell{57.2 (-2.3)} & \makecell{64.6 (-1.3)} & \makecell{53.1 (-0.5)} & \makecell{47.9 (-0.2)} \\
\textbf{Qwen3-8B-Base}  & \makecell{68.1 (+7.6)} & \makecell{95.6 (+10.1)} & \makecell{73.6 (+0.0)} & \makecell{72.3 (-0.4)} & \makecell{62.3 (+5.1)} & \makecell{48.6 (+0.6)} \\
\textbf{Qwen3-8B}  & \makecell{52.7 (+5.0)} & \makecell{92.8 (+18.0)} & \makecell{67.1 (-1.7)} & \makecell{70.1 (-0.7)} & \makecell{60.5 (-0.7)} & \makecell{48.6 (-0.3)} \\
\textbf{Qwen3-14B-Base}  & \makecell{77.1 (+11.3)} & \makecell{95.1 (+6.5)} & \makecell{76.1 (-0.2)} & \makecell{70.4 (+1.4)} & \makecell{73.4 (+5.7)} & \makecell{51.6 (+0.8)} \\
\textbf{Qwen3-14B}  & \makecell{51.6 (+3.7)} & \makecell{94.2 (+3.6)} & \makecell{75.6 (+0.1)} & \makecell{70.8 (+0.6)} & \makecell{64.2 (+0.2)} & \makecell{54.0 (-0.1)} \\
\midrule

\textbf{Llama-3.2-1B}  & \makecell{38.6 (+0.2)} & \makecell{28.4 (+2.2)} & \makecell{50.2 (-1.0)} & \makecell{50.2 (+0.5)} & \makecell{49.2 (-0.2)} & \makecell{48.6 (-1.2)} \\
\textbf{Llama-3.2-1B-Instruct}  & \makecell{37.3 (+0.1)} & \makecell{80.5 (+46.9)} & \makecell{50.2 (-0.4)} & \makecell{51.9 (-0.5)} & \makecell{50.4 (+0.0)} & \makecell{48.0 (-0.1)} \\
\textbf{Llama-3.2-3B}  & \makecell{45.4 (+0.2)} & \makecell{68.1 (+16.1)} & \makecell{57.3 (+0.6)} & \makecell{55.2 (+1.0)} & \makecell{55.3 (+1.1)} & \makecell{47.6 (+0.2)} \\
\textbf{Llama-3.2-3B-Instruct}  & \makecell{49.4 (+8.3)} & \makecell{93.8 (+32.0)} & \makecell{59.3 (-0.2)} & \makecell{57.8 (+0.1)} & \makecell{51.6 (+0.1)} & \makecell{48.2 (+0.3)} \\
\textbf{Llama-3.1-8B}  & \makecell{46.5 (+6.1)} & \makecell{95.2 (+41.6)} & \makecell{61.2 (+0.9)} & \makecell{61.5 (+0.6)} & \makecell{54.4 (+1.4)} & \makecell{48.4 (+0.2)} \\
\textbf{Llama-3.1-8B-Instruct}  & \makecell{50.5 (+13.1)} & \makecell{97.1 (+39.9)} & \makecell{61.4 (+5.1)} & \makecell{60.5 (+2.8)} & \makecell{64.2 (+10.1)} & \makecell{48.8 (+0.7)} \\
\midrule

\textbf{Mistral-7B-v0.3}  & \makecell{57.9 (+6.4)} & \makecell{99.4 (+52.5)} & \makecell{54.1 (+1.7)} & \makecell{55.6 (-0.7)} & \makecell{73.3 (+15.3)} & \makecell{53.4 (+5.3)} \\
\textbf{Mistral-7B-Instruct-v0.3}  & \makecell{69.2 (+3.4)} & \makecell{99.7 (+26.7)} & \makecell{58.7 (+0.4)} & \makecell{59.1 (+0.0)} & \makecell{79.6 (+2.5)} & \makecell{62.0 (+6.0)} \\
\textbf{Mistral-Nemo-Base-2407 (12B)}  & \makecell{60.0 (+10.3)} & \makecell{97.4 (+46.2)} & \makecell{57.2 (+2.6)} & \makecell{65.4 (+0.1)} & \makecell{71.2 (+4.7)} & \makecell{46.4 (-0.5)} \\
\textbf{Mistral-Nemo-Instruct-2407 (12B)}  & \makecell{67.2 (+11.6)} & \makecell{98.6 (+33.5)} & \makecell{64.2 (+2.0)} & \makecell{65.4 (+2.3)} & \makecell{75.1 (-1.0)} & \makecell{47.8 (+0.2)} \\
\midrule

\textbf{Mean $\Delta$ over Baseline} & 6.4 & 30.7 & 0.7 & 0.1 & 1.8 & 0.3 \\

\bottomrule
\end{tabular}}
\caption{Model performance on \bench, KMMLU, and KoBest BoolQ after fine-tuning targeting the symbol format. Values in parentheses indicate the performance gain relative to the baseline results.}
\label{tab:sft_symbol1}
\end{table*}

\begin{table*}[htbp]
\centering
\footnotesize
\renewcommand{\arraystretch}{1.3}
\resizebox{\textwidth}{!}{%
\begin{tabular}{lcccc}
\toprule

 & \multicolumn{2}{c}{\textbf{ARC}} & \textbf{Hellaswag} & \textbf{Winogrande} \\
\cmidrule(l){2-3}
\textbf{Model} & \textbf{easy} & \textbf{challenge} & & \\
 \midrule \midrule

\textbf{Midm-2.0-Mini-Instruct (2.3B)}  & \makecell{79.4 (+0.9)} & \makecell{50.8 (+0.4)} & \makecell{70.3 (+0.1)} & \makecell{67.1 (+0.3)} \\
\textbf{Midm-2.0-Base-Instruct (11.5B)}  & \makecell{84.9 (+0.4)} & \makecell{61.8 (+0.3)} & \makecell{80.4 (+0.0)} & \makecell{73.8 (+0.4)} \\
\midrule

\textbf{EXAONE-4.0-1.2B}  & \makecell{33.2 (+0.2)} & \makecell{24.8 (+0.1)} & \makecell{32.1 (+1.0)} & \makecell{49.3 (-0.8)} \\
\textbf{EXAONE-Deep-2.4B}  & \makecell{56.3 (+2.0)} & \makecell{37.5 (-0.9)} & \makecell{55.7 (+0.5)} & \makecell{54.5 (-0.2)} \\
\textbf{EXAONE-Deep-7.8B}  & \makecell{65.2 (+0.2)} & \makecell{44.8 (+1.0)} & \makecell{63.6 (+0.0)} & \makecell{57.3 (-0.5)} \\
\midrule

\textbf{kanana-1.5-2.1b-base}  & \makecell{78.5 (+0.2)} & \makecell{51.0 (+0.4)} & \makecell{66.0 (+0.0)} & \makecell{64.6 (+0.7)} \\
\textbf{kanana-1.5-2.1b-instruct-2505}  & \makecell{74.7 (+0.8)} & \makecell{49.9 (-0.2)} & \makecell{66.7 (-0.2)} & \makecell{62.5 (-0.5)} \\
\textbf{kanana-1.5-8b-base}  & \makecell{82.6 (+1.3)} & \makecell{56.6 (+2.5)} & \makecell{77.3 (-0.1)} & \makecell{72.6 (-0.4)} \\
\textbf{kanana-1.5-8b-instruct-2505}  & \makecell{34.2 (+0.1)} & \makecell{29.9 (+0.0)} & \makecell{78.8 (+0.1)} & \makecell{71.6 (-0.2)} \\
\midrule

\textbf{HyperCLOVAX-SEED-Text-Instruct-0.5B}  & \makecell{65.2 (-0.2)} & \makecell{38.1 (+0.8)} & \makecell{52.3 (+0.1)} & \makecell{55.2 (+0.6)} \\
\textbf{HyperCLOVAX-SEED-Text-Instruct-1.5B}  & \makecell{66.1 (+0.0)} & \makecell{44.1 (+0.2)} & \makecell{60.7 (-0.2)} & \makecell{56.9 (-0.5)} \\
\textbf{HyperCLOVAX-SEED-Think-14B}  & \makecell{75.2 (+0.3)} & \makecell{54.2 (-0.5)} & \makecell{79.4 (+0.0)} & \makecell{71.5 (+0.3)} \\
\midrule

\textbf{A.X-4.0-Light (7.2B)}  & \makecell{75.0 (+0.5)} & \makecell{56.2 (+0.3)} & \makecell{77.2 (+0.1)} & \makecell{71.6 (+0.5)} \\
\midrule

\textbf{SOLAR-10.7B-v1.0}  & \makecell{79.0 (+1.0)} & \makecell{55.9 (+0.4)} & \makecell{83.2 (+0.1)} & \makecell{75.5 (+0.8)} \\
\textbf{SOLAR-10.7B-Instruct-v1.0}  & \makecell{80.5 (-0.6)} & \makecell{61.7 (-0.5)} & \makecell{86.8 (+0.2)} & \makecell{76.2 (+1.1)} \\
\midrule

\textbf{Qwen3-0.6B-Base}  & \makecell{61.0 (+3.1)} & \makecell{38.5 (+0.5)} & \makecell{53.7 (+0.1)} & \makecell{58.1 (-1.0)} \\
\textbf{Qwen3-0.6B}  & \makecell{31.6 (+0.2)} & \makecell{28.0 (-0.2)} & \makecell{47.3 (+0.1)} & \makecell{56.0 (+0.3)} \\
\textbf{Qwen3-1.7B-Base}  & \makecell{70.2 (+1.9)} & \makecell{45.5 (+0.6)} & \makecell{66.6 (+0.1)} & \makecell{64.8 (+0.4)} \\
\textbf{Qwen3-1.7B}  & \makecell{69.3 (-0.4)} & \makecell{43.0 (+0.2)} & \makecell{60.2 (-0.2)} & \makecell{61.6 (-0.1)} \\
\textbf{Qwen3-4B-Base}  & \makecell{77.7 (+1.9)} & \makecell{53.1 (+1.6)} & \makecell{73.7 (+0.0)} & \makecell{70.6 (+0.1)} \\
\textbf{Qwen3-4B}  & \makecell{76.9 (-1.6)} & \makecell{52.3 (-1.4)} & \makecell{68.4 (+0.0)} & \makecell{66.0 (+0.1)} \\
\textbf{Qwen3-8B-Base}  & \makecell{81.1 (+0.8)} & \makecell{57.8 (+1.1)} & \makecell{78.7 (+0.1)} & \makecell{72.4 (-0.3)} \\
\textbf{Qwen3-8B}  & \makecell{80.3 (-0.6)} & \makecell{56.1 (-0.2)} & \makecell{74.9 (+0.0)} & \makecell{68.1 (+0.5)} \\
\textbf{Qwen3-14B-Base}  & \makecell{82.6 (+0.6)} & \makecell{59.6 (+0.6)} & \makecell{81.4 (+0.0)} & \makecell{74.1 (+0.1)} \\
\textbf{Qwen3-14B}  & \makecell{83.1 (+0.3)} & \makecell{60.9 (+0.7)} & \makecell{78.8 (-0.1)} & \makecell{73.3 (+0.3)} \\
\midrule

\textbf{Llama-3.2-1B}  & \makecell{61.9 (+0.1)} & \makecell{36.8 (-0.3)} & \makecell{64.2 (+0.1)} & \makecell{60.8 (+0.1)} \\
\textbf{Llama-3.2-1B-Instruct}  & \makecell{63.2 (-0.6)} & \makecell{37.5 (-0.2)} & \makecell{61.3 (-0.3)} & \makecell{61.4 (+0.1)} \\
\textbf{Llama-3.2-3B}  & \makecell{72.0 (+0.2)} & \makecell{46.3 (+0.1)} & \makecell{74.2 (+0.1)} & \makecell{69.2 (-0.5)} \\
\textbf{Llama-3.2-3B-Instruct}  & \makecell{71.3 (+0.2)} & \makecell{46.2 (+0.0)} & \makecell{71.6 (+0.0)} & \makecell{68.6 (-0.2)} \\
\textbf{Llama-3.1-8B}  & \makecell{82.3 (-0.3)} & \makecell{55.3 (+0.5)} & \makecell{79.3 (+0.0)} & \makecell{74.4 (+0.0)} \\
\textbf{Llama-3.1-8B-Instruct}  & \makecell{77.3 (+0.1)} & \makecell{54.1 (-0.1)} & \makecell{79.5 (-0.2)} & \makecell{74.3 (+0.2)} \\
\midrule

\textbf{Mistral-7B-v0.3}  & \makecell{80.5 (+0.4)} & \makecell{54.9 (+0.5)} & \makecell{80.7 (+0.1)} & \makecell{74.4 (+0.4)} \\
\textbf{Mistral-7B-Instruct-v0.3}  & \makecell{81.1 (-0.7)} & \makecell{59.8 (-0.4)} & \makecell{83.4 (+0.1)} & \makecell{74.8 (+0.0)} \\
\textbf{Mistral-Nemo-Base-2407 (12B)}  & \makecell{83.6 (+0.1)} & \makecell{60.0 (+0.4)} & \makecell{83.0 (+0.0)} & \makecell{76.5 (-0.4)} \\
\textbf{Mistral-Nemo-Instruct-2407 (12B)}  & \makecell{80.6 (+0.0)} & \makecell{59.6 (+0.0)} & \makecell{82.5 (+0.1)} & \makecell{76.6 (-0.3)} \\
\midrule

\textbf{Mean $\Delta$ over Baseline} & 0.4 & 0.2 & 0.0 & 0.0 \\
\bottomrule
\end{tabular}%
}
\caption{Model performance on ARC, Hellaswag, and Winogrande after fine-tuning targeting the symbol format. Values in parentheses indicate the performance gain relative to the baseline results.}
\label{tab:sft_symbol2}
\end{table*}

%% file: tables/app_error_analysis.tex
\begin{table*}[htbp]
\centering
\footnotesize
\resizebox{\textwidth}{!}{%
\begin{tabular}{@{}lcc|ccc@{}}
\toprule
\textbf{model name} & \makecell{\textbf{evaluation}\\\textbf{method}} & \textbf{performance} & \multicolumn{3}{c@{}}{\textbf{incorrect  choice distribution}} \\
\cmidrule(l){4-6}
& & & \textbf{local negation(\%)} & \textbf{contradiction(\%)} & \textbf{paraphrase(\%)} \\
\midrule
\multicolumn{1}{l}{\multirow{2}{*}{\textbf{Qwen3-0.6B-Base}}} & \textbf{cloze} & 38.94 & 95.15 & 3.32 & 1.53 \\
\addlinespace[2pt]
\multicolumn{1}{l}{} & \textbf{symbol} & 51.56 & 61.09 & 12.38 & 26.53 \\
\addlinespace[2pt]

\multicolumn{1}{l}{\multirow{2}{*}{\textbf{Qwen3-0.6B}}} & \textbf{cloze} & 36.92 & 94.94 & 3.58 & 1.48 \\
\addlinespace[2pt]
\multicolumn{1}{l}{} & \textbf{symbol} & 64.56 & 75.16 & 7.25 & 17.58 \\

\multicolumn{1}{l}{\multirow{2}{*}{\textbf{Qwen3-1.7B-Base}}} & \textbf{cloze} & 60.28 & 94.31 & 4.51 & 1.18 \\
\addlinespace[2pt]
\multicolumn{1}{l}{} & \textbf{symbol} & 64.64 & 66.30 & 7.71 & 25.99 \\
\addlinespace[2pt]

\multicolumn{1}{l}{\multirow{2}{*}{\textbf{Qwen3-1.7B}}} & \textbf{cloze} & 51.79 & 93.05 & 5.17 & 1.78 \\
\addlinespace[2pt]
\multicolumn{1}{l}{} & \textbf{symbol} & 76.01 & 71.75 & 7.14 & 21.10 \\
\addlinespace[2pt]

\multicolumn{1}{l}{\multirow{2}{*}{\textbf{Qwen3-4B-Base}}} & \textbf{cloze} & 64.88 & 93.35 & 4.66 & 2.00 \\
\addlinespace[2pt]
\multicolumn{1}{l}{} & \textbf{symbol} & 82.17 & 82.53 & 12.66 & 4.80 \\
\addlinespace[2pt]

\multicolumn{1}{l}{\multirow{2}{*}{\textbf{Qwen3-4B}}} & \textbf{cloze} & 45.56 & 95.14 & 3.58 & 1.29 \\
\addlinespace[2pt]
\multicolumn{1}{l}{} & \textbf{symbol} & 80.45 & 86.85 & 7.57 & 5.58 \\
\addlinespace[2pt]

\multicolumn{1}{l}{\multirow{2}{*}{\textbf{Qwen3-8B-Base}}} & \textbf{cloze} & 56.31 & 94.83 & 3.74 & 1.43 \\
\addlinespace[2pt]
\multicolumn{1}{l}{} & \textbf{symbol} & 85.44 & 85.56 & 8.02 & 6.42 \\
\addlinespace[2pt]

\multicolumn{1}{l}{\multirow{2}{*}{\textbf{Qwen3-8B}}} & \textbf{cloze} & 46.11 & 93.64 & 4.48 & 1.88 \\
\addlinespace[2pt]
\multicolumn{1}{l}{} & \textbf{symbol} & 74.77 & 86.73 & 7.72 & 5.56 \\
\addlinespace[2pt]

\multicolumn{1}{l}{\multirow{2}{*}{\textbf{Qwen3-14B-Base}}} & \textbf{cloze} & 62.31 & 94.21 & 4.13 & 1.65 \\
\addlinespace[2pt]
\multicolumn{1}{l}{} & \textbf{symbol} & 88.55 & 74.83 & 18.37 & 6.80 \\
\addlinespace[2pt]

\multicolumn{1}{l}{\multirow{2}{*}{\textbf{Qwen3-14B}}} & \textbf{cloze} & 46.18 & 94.21 & 4.34 & 1.45 \\
\addlinespace[2pt]
\multicolumn{1}{l}{} & \textbf{symbol} & 90.50 & 77.05 & 18.03 & 4.92 \\
\addlinespace[2pt]

\multicolumn{1}{l}{\multirow{2}{*}{\textbf{Qwen3-32B}}} & \textbf{cloze} & 54.21 & 94.56 & 3.91 & 1.53 \\
\addlinespace[2pt]
\multicolumn{1}{l}{} & \textbf{symbol} & 94.16 & 89.33 & 6.67 & 4.00 \\
\midrule

\multicolumn{1}{l}{\multirow{2}{*}{\textbf{Llama-3.2-1B}}} & \textbf{cloze} & 37.38 & 93.91 & 4.35 & 1.74 \\
\addlinespace[2pt]
\multicolumn{1}{l}{} & \textbf{symbol} & 26.17 & 35.34 & 31.01 & 33.65 \\
\addlinespace[2pt]

\multicolumn{1}{l}{\multirow{2}{*}{\textbf{Llama-3.2-1B-Instruct}}} & \textbf{cloze} & 35.59 & 91.29 & 6.65 & 2.06 \\
\addlinespace[2pt]
\multicolumn{1}{l}{} & \textbf{symbol} & 33.49 & 38.88 & 18.97 & 42.15 \\
\addlinespace[2pt]

\multicolumn{1}{l}{\multirow{2}{*}{\textbf{Llama-3.2-3B}}} & \textbf{cloze} & 43.46 & 94.63 & 3.31 & 2.07 \\
\addlinespace[2pt]
\multicolumn{1}{l}{} & \textbf{symbol} & 51.95 & 74.39 & 11.02 & 14.59 \\
\addlinespace[2pt]

\multicolumn{1}{l}{\multirow{2}{*}{\textbf{Llama-3.2-3B-Instruct}}} & \textbf{cloze} & 38.40 & 91.66 & 6.07 & 2.28 \\
\addlinespace[2pt]
\multicolumn{1}{l}{} & \textbf{symbol} & 61.68 & 77.64 & 9.55 & 12.80 \\
\addlinespace[2pt]

\multicolumn{1}{l}{\multirow{2}{*}{\textbf{Llama-3.1-8B}}} & \textbf{cloze} & 39.56 & 93.17 & 4.51 & 2.32 \\
\addlinespace[2pt]
\multicolumn{1}{l}{} & \textbf{symbol} & 53.66 & 53.28 & 12.27 & 34.45 \\
\addlinespace[2pt]

\multicolumn{1}{l}{\multirow{2}{*}{\textbf{Llama-3.1-8B-Instruct}}} & \textbf{cloze} & 35.98 & 91.61 & 6.08 & 2.31 \\
\addlinespace[2pt]
\multicolumn{1}{l}{} & \textbf{symbol} & 57.09 & 66.24 & 10.34 & 23.41 \\
\addlinespace[2pt]

\multicolumn{1}{l}{\multirow{2}{*}{\textbf{Llama-3.1-70B}}} & \textbf{cloze} & 53.97 & 95.43 & 3.55 & 1.02 \\
\addlinespace[2pt]
\multicolumn{1}{l}{} & \textbf{symbol} & 76.56 & 77.41 & 8.97 & 13.62 \\
\addlinespace[2pt]

\multicolumn{1}{l}{\multirow{2}{*}{\textbf{Llama-3.1-70B-Instruct}}} & \textbf{cloze} & 48.05 & 94.90 & 4.20 & 0.90 \\
\addlinespace[2pt]
\multicolumn{1}{l}{} & \textbf{symbol} & 80.92 & 87.76 & 6.12 & 6.12 \\
\addlinespace[2pt]

\bottomrule
\end{tabular}%
}
\caption{Incorrect choice distributions for the \textbf{Qwen3} and \textbf{Llama} family under zero-shot conditions.}
\label{tab:error_analysis_qwen_llama}
\end{table*}


\begin{table*}[htbp]
\centering
\footnotesize
\resizebox{\textwidth}{!}{%
\begin{tabular}{@{}lcc|ccc@{}}
\toprule
\textbf{model name} & \makecell{\textbf{evaluation}\\\textbf{method}} & \textbf{performance} & \multicolumn{3}{c@{}}{\textbf{incorrect  choice distribution}} \\
\cmidrule(l){4-6}
& & & \textbf{local negation(\%)} & \textbf{contradiction(\%)} & \textbf{paraphrase(\%)} \\
\midrule

\multicolumn{1}{l}{\multirow{2}{*}{\textbf{Mistral-7B-v0.3}}} & \textbf{cloze} & 49.14 & 94.18 & 4.44 & 1.38 \\
\addlinespace[2pt]
\multicolumn{1}{l}{} & \textbf{symbol} & 46.81 & 36.75 & 14.35 & 48.90 \\
\addlinespace[2pt]

\multicolumn{1}{l}{\multirow{2}{*}{\textbf{Mistral-7B-Instruct-v0.3}}} & \textbf{cloze} & 63.16 & 93.66 & 4.65 & 1.69 \\
\addlinespace[2pt]
\multicolumn{1}{l}{} & \textbf{symbol} & 72.90 & 75.86 & 6.03 & 18.10 \\
\addlinespace[2pt]

\multicolumn{1}{l}{\multirow{2}{*}{\textbf{Mistral-Nemo-Base-2407 (12B)}}} & \textbf{cloze} & 47.35 & 93.34 & 5.18 & 1.48 \\
\addlinespace[2pt]
\multicolumn{1}{l}{} & \textbf{symbol} & 51.17 & 63.32 & 10.85 & 25.84 \\
\addlinespace[2pt]

\multicolumn{1}{l}{\multirow{2}{*}{\textbf{Mistral-Nemo-Instruct-2407 (12B)}}} & \textbf{cloze} & 52.57 & 92.12 & 5.91 & 1.97 \\
\addlinespace[2pt]
\multicolumn{1}{l}{} & \textbf{symbol} & 65.03 & 92.65 & 4.90 & 2.45 \\
\addlinespace[2pt]

\multicolumn{1}{l}{\multirow{2}{*}{\textbf{Mistral-Small-24B-Base-2501}}} & \textbf{cloze} & 57.40 & 94.15 & 4.57 & 1.28 \\
\addlinespace[2pt]
\multicolumn{1}{l}{} & \textbf{symbol} & 71.50 & 63.93 & 23.50 & 12.57 \\
\addlinespace[2pt]

\multicolumn{1}{l}{\multirow{2}{*}{\textbf{Mistral-Small-24B-Instruct-2501}}} & \textbf{cloze} & 56.93 & 96.38 & 3.07 & 0.54 \\
\addlinespace[2pt]
\multicolumn{1}{l}{} & \textbf{symbol} & 81.78 & 81.62 & 14.53 & 3.85 \\
\midrule

\multicolumn{1}{l}{\textbf{gpt-4.1-mini}} & \textbf{symbol} & 83.49 & 91.98 & 6.60 & 1.42 \\
\addlinespace[2pt]

\multicolumn{1}{l}{\textbf{gpt-4.1}} & \textbf{symbol} & 92.13 & 97.03 & 1.98 & 0.99 \\
\addlinespace[2pt]
\midrule

\multicolumn{1}{l}{\textbf{claude-haiku-4-5-20251001}} & \textbf{symbol} & 92.76 & 84.95 & 11.83 & 3.23 \\
\addlinespace[2pt]

\multicolumn{1}{l}{\textbf{claude-sonnet-4-5-20250929}} & \textbf{symbol} & 98.05 & 72.00 & 28.00 & 0.00 \\

\bottomrule
\end{tabular}%
}
\caption{Incorrect choice distributions for the \textbf{Mistral}, \textbf{GPT-4.1}, and \textbf{Claude-Opus 4.5} family under zero-shot conditions.}
\label{tab:error_analysis_mistral_gpt_claude}
\end{table*}

\begin{table*}[htbp]
\centering
\footnotesize
\resizebox{\textwidth}{!}{%
\begin{tabular}{@{}lcc|ccc@{}}
\toprule
\textbf{model name} & \makecell{\textbf{evaluation}\\\textbf{method}} & \textbf{performance} & \multicolumn{3}{c@{}}{\textbf{incorrect  choice distribution}} \\
\cmidrule(l){4-6}
& & & \textbf{local negation(\%)} & \textbf{contradiction(\%)} & \textbf{paraphrase(\%)} \\
\midrule


\multicolumn{1}{l}{\multirow{2}{*}{\textbf{Midm-2.0-Mini-Instruct (2.3B)}}} & \textbf{cloze} & 56.00 & 88.32 & 9.03 & 2.65 \\
\addlinespace[2pt]
\multicolumn{1}{l}{} & \textbf{symbol} & 55.30 & 58.71 & 12.89 & 28.40 \\
\addlinespace[2pt]
\multicolumn{1}{l}{\multirow{2}{*}{\textbf{Midm-2.0-Base-Instruct (11.5B)}}} & \textbf{cloze} & 61.21 & 94.38 & 4.42 & 1.20 \\
\addlinespace[2pt]
\multicolumn{1}{l}{} & \textbf{symbol} & 59.50 & 50.77 & 30.77 & 18.46 \\
\midrule

\multicolumn{1}{l}{\multirow{2}{*}{\textbf{EXAONE-Deep-2.4B}}} & \textbf{cloze} & 35.75 & 89.45 & 7.64 & 2.91 \\
\addlinespace[2pt]
\multicolumn{1}{l}{} & \textbf{symbol} & 40.19 & 50.65 & 13.28 & 36.07 \\
\addlinespace[2pt]

\multicolumn{1}{l}{\multirow{2}{*}{\textbf{EXAONE-Deep-7.8B}}} & \textbf{cloze} & 56.00 & 93.45 & 5.13 & 1.42 \\
\addlinespace[2pt]
\multicolumn{1}{l}{} & \textbf{symbol} & 43.61 & 42.40 & 17.40 & 40.19 \\
\addlinespace[2pt]

\multicolumn{1}{l}{\multirow{2}{*}{\textbf{EXAONE-Deep-32B}}} & \textbf{cloze} & 58.33 & 93.27 & 4.86 & 1.87 \\
\addlinespace[2pt]
\multicolumn{1}{l}{} & \textbf{symbol} & 71.65 & 76.92 & 9.34 & 13.74 \\
\addlinespace[2pt]

\multicolumn{1}{l}{\multirow{2}{*}{\textbf{EXAONE-4.0-1.2B}}} & \textbf{cloze} & 28.82 & 86.21 & 10.28 & 3.50 \\
\addlinespace[2pt]
\multicolumn{1}{l}{} & \textbf{symbol} & 26.48 & 36.65 & 31.04 & 32.31 \\
\addlinespace[2pt]
\multicolumn{1}{l}{\multirow{2}{*}{\textbf{EXAONE-4.0-32B}}} & \textbf{cloze} & 50.16 & 96.41 & 3.12 & 0.47 \\
\addlinespace[2pt]
\multicolumn{1}{l}{} & \textbf{symbol} & 81.46 & 90.76 & 5.88 & 3.36 \\

\midrule

\multicolumn{1}{l}{\multirow{2}{*}{\textbf{kanana-1.5-2.1b-base}}} & \textbf{cloze} & 47.98 & 94.31 & 3.59 & 2.10 \\
\addlinespace[2pt]
\multicolumn{1}{l}{} & \textbf{symbol} & 45.09 & 42.98 & 12.06 & 44.96 \\
\addlinespace[2pt]

\multicolumn{1}{l}{\multirow{2}{*}{\textbf{kanana-1.5-2.1b-instruct-2505}}} & \textbf{cloze} & 61.99 & 91.80 & 6.35 & 1.84 \\
\addlinespace[2pt]
\multicolumn{1}{l}{} & \textbf{symbol} & 57.40 & 63.62 & 10.97 & 25.41 \\
\addlinespace[2pt]

\multicolumn{1}{l}{\multirow{2}{*}{\textbf{kanana-1.5-8b-base}}} & \textbf{cloze} & 43.54 & 96.55 & 2.07 & 1.38 \\
\addlinespace[2pt]
\multicolumn{1}{l}{} & \textbf{symbol} & 43.54 & 31.31 & 17.52 & 51.17 \\
\addlinespace[2pt]

\multicolumn{1}{l}{\multirow{2}{*}{\textbf{kanana-1.5-8b-instruct-2505}}} & \textbf{cloze} & 53.50 & 94.64 & 4.19 & 1.17 \\
\addlinespace[2pt]
\multicolumn{1}{l}{} & \textbf{symbol} & 56.70 & 51.44 & 13.67 & 34.89 \\
\midrule

\multicolumn{1}{l}{\multirow{2}{*}{\textbf{HyperCLOVAX-SEED-Text-Instruct-0.5B}}} & \textbf{cloze} & 35.12 & 93.88 & 3.84 & 2.28 \\
\addlinespace[2pt]
\multicolumn{1}{l}{} & \textbf{symbol} & 52.65 & 62.17 & 10.86 & 26.97 \\
\addlinespace[2pt]

\multicolumn{1}{l}{\multirow{2}{*}{\textbf{HyperCLOVAX-SEED-Text-Instruct-1.5B}}} & \textbf{cloze} & 36.14 & 93.05 & 5.24 & 1.71 \\
\addlinespace[2pt]
\multicolumn{1}{l}{} & \textbf{symbol} & 46.57 & 47.23 & 23.47 & 29.30 \\
\addlinespace[2pt]

\multicolumn{1}{l}{\multirow{2}{*}{\textbf{HyperCLOVAX-SEED-Think-14B}}} & \textbf{cloze} & 45.56 & 98.00 & 1.43 & 0.57 \\
\addlinespace[2pt]
\multicolumn{1}{l}{} & \textbf{symbol} & 78.04 & 70.57 & 14.89 & 14.54 \\

\midrule


\multicolumn{1}{l}{\multirow{2}{*}{\textbf{A.X-4.0-Light (7.2B)}}} & \textbf{cloze} & 69.08 & 88.41 & 8.31 & 3.27 \\
\addlinespace[2pt]
\multicolumn{1}{l}{} & \textbf{symbol} & 84.89 & 84.54 & 5.15 & 10.31 \\
\addlinespace[2pt]
\multicolumn{1}{l}{\multirow{2}{*}{\textbf{A.X-4.0 (72B)}}} & \textbf{cloze} & 73.83 & 91.67 & 5.65 & 2.68 \\
\addlinespace[2pt]
\multicolumn{1}{l}{} & \textbf{symbol} & 95.72 & 87.27 & 7.27 & 5.45 \\
\midrule


\multicolumn{1}{l}{\multirow{2}{*}{\textbf{SOLAR-10.7B-v1.0}}} & \textbf{cloze} & 40.50 & 93.98 & 4.71 & 1.31 \\
\addlinespace[2pt]
\multicolumn{1}{l}{} & \textbf{symbol} & 41.98 & 44.83 & 14.23 & 40.94 \\
\addlinespace[2pt]
\multicolumn{1}{l}{\multirow{2}{*}{\textbf{SOLAR-10.7B-Instruct-v1.0}}} & \textbf{cloze} & 61.29 & 92.96 & 5.03 & 2.01 \\
\addlinespace[2pt]
\multicolumn{1}{l}{} & \textbf{symbol} & 54.67 & 77.84 & 11.68 & 10.48 \\

\bottomrule
\end{tabular}%
}
\caption{Incorrect choice distributions for the Korean Models (\textbf{mi:dm 2.0}, \textbf{EXAONE}, \textbf{Kanana 1.5}, \textbf{HyperCLOVAX}, \textbf{A.X 4.0}, and \textbf{SOLAR} model family) under zero-shot conditions.}
\label{tab:error_analysis_korean}
\end{table*}